\documentclass{article}
\usepackage[T1]{fontenc}
\usepackage{xcolor}

\usepackage[final]{corl_2026}

\usepackage{caption}
\usepackage{capt-of}
\usepackage{multicol}
\usepackage{placeins}
\usepackage{graphicx}
\usepackage{amsmath}
\usepackage{amsfonts}
\usepackage{booktabs}
\usepackage[font=normalsize]{subfig}

\definecolor{darkgreen}{rgb}{0.0,0.5,0.0}

\newcommand{\ourmethod}{FPO++}
\newcommand{\revision}[1]{#1}

\title{Flow Policy Gradients for Robot Control}

\author{
  \mdseries Brent Yi$^{1,2,\dagger,*}$, Hongsuk Choi$^{1,2,\dagger,*}$, Himanshu Gaurav Singh$^{1,2,\dagger}$, Xiaoyu Huang$^{1,2,\dagger}$ \\
  \mdseries Takara E.~Truong$^{1,3,\dagger}$, Carmelo Sferrazza$^{1}$, Yi Ma$^{2,4}$, Rocky Duan$^{1,\ddagger}$ \\
  \mdseries Pieter Abbeel$^{1,2,\ddagger}$, Guanya Shi$^{1,5,\ddagger}$, Karen Liu$^{1,3,\ddagger}$, Angjoo Kanazawa$^{1,2,\ddagger}$ \\
  $^1$Amazon FAR \quad $^2$UC Berkeley \quad $^3$Stanford \quad $^4$HKU \quad $^5$CMU \\
  {\footnotesize *Equal Contribution \quad $\dagger$Work done as an intern at Amazon FAR \quad $\ddagger$Amazon FAR team co-lead}
}

\begin{document}
\hypersetup{pdfauthor={Brent Yi, Hongsuk Choi, Himanshu Gaurav Singh, Xiaoyu Huang, Takara E. Truong, Carmelo Sferrazza, Yi Ma, Rocky Duan, Pieter Abbeel, Guanya Shi, Karen Liu, Angjoo Kanazawa}}
\maketitle

\vspace{-1.25em}
\begin{figure}[h]
    \centering
    \includegraphics[width=1.0\textwidth]{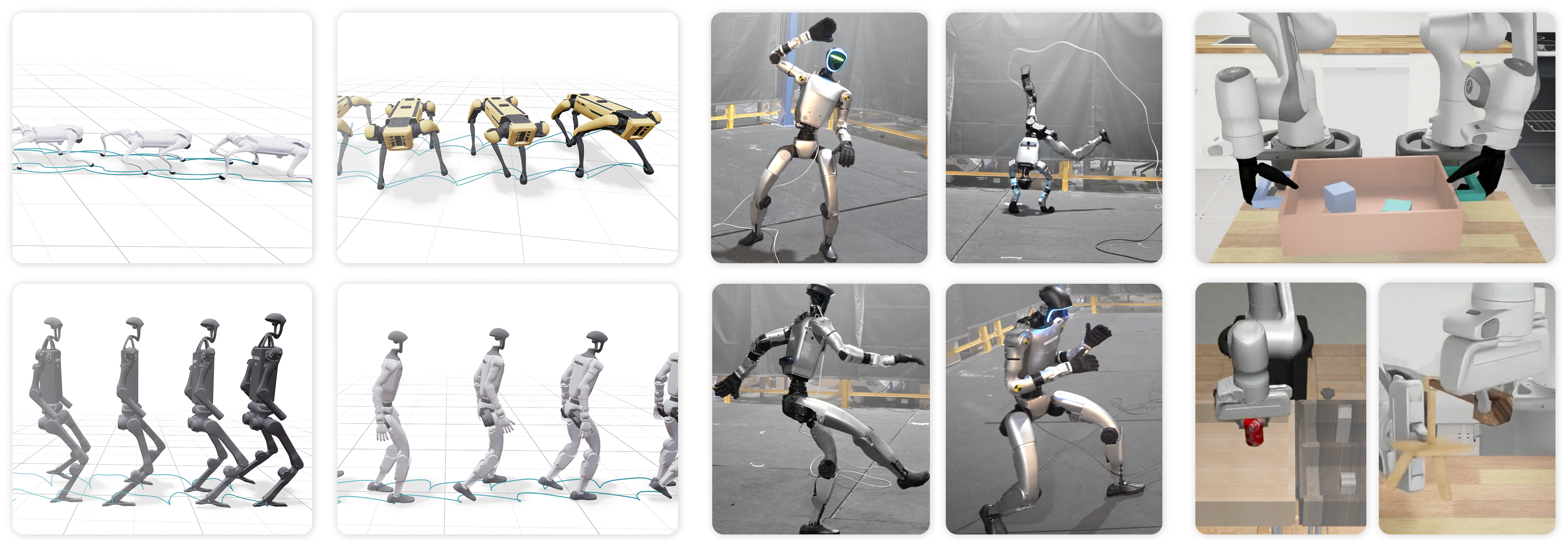}
    \caption{
        \textbf{Flow policies for robot control.}
        We show how robust control policies for quadrupeds, humanoids, and manipulators can be trained with reinforcement learning.
    }
    \label{fig:splash}
\end{figure}

\begin{abstract}
Likelihood-based policy gradient methods are the dominant approach for training robot control policies from rewards. These methods rely on differentiable action likelihoods, which constrain policy outputs to simple distributions like Gaussians. In this work, we show how flow matching policy gradients, a recent framework that bypasses likelihood computation, can be made effective for training and \revision{finetuning} more expressive policies in challenging robot control settings. We introduce an improved objective that enables success in legged locomotion, humanoid motion tracking, and manipulation tasks, as well as robust sim-to-real transfer on two humanoid robots. We then present ablations and analysis on training dynamics. Results show how policies can exploit the flow representation for exploration when training from scratch, as well as improved \revision{finetuning} robustness over baselines.
{\hypersetup{urlcolor=blue}Project webpage: \href{https://hongsukchoi.github.io/fpo-control/}{hongsukchoi.github.io/fpo-control}.}
\end{abstract}


\section{Introduction}
\label{sec:introduction}

Likelihood-based policy gradient methods~\citep{sutton1999policy,schulman2017proximal} have driven a broad range of robot control results.
This includes recent advances in legged locomotion~\citep{rudin2022learning}, whole-body humanoid control~\citep{liao2026beyondmimic,videomimic}, and in-hand manipulation~\citep{qi2023general}, where policies are trained from scratch, as well as for manipulation policies that are pretrained with demonstrations~\citep{ren2024diffusion,zhang2025reinflow}.

Policy gradient algorithms are traditionally run by sampling actions from stochastic policies, followed by policy updates that backpropagate through the differentiable likelihoods of sampled actions~\citep{sutton1999policy}.
This is effective for simple action distributions, but challenging for more expressive policy representations: computing likelihoods in the flow policies used for imitation learning~\citep{chi2025diffusion,black2024pi0visionlanguageactionflowmodel}, for example, requires expensive sampling or integration to account for volume changes in the underlying flow field~\citep{skreta2025superpositiondiffusionmodelsusing,grathwohl2018ffjord}.
Reinforcement learning (RL) algorithms that support these representations promise more expressive distributions for exploration when training from scratch~\citep{mcallister2025flow}, as well as post-training of more capable pretrained policies~\citep{ren2024diffusion,zhang2025reinflow}---ideally within a single algorithm and policy representation that spans both regimes.

We study robot control with flow matching policy gradients~\citep{mcallister2025flow}, a framework for flow RL that uses conditional flow matching to circumvent likelihoods entirely.
Specifically, we show how this class of algorithm---previously only validated in simpler synthetic settings---can be made practical for training robot control policies in more challenging settings.

Our contributions are as follows.
\textbf{(1)}~We introduce an improved flow policy gradient algorithm that we call FPO++.
FPO++ proposes two simple changes, per-sample ratio clipping and an asymmetric trust region, that enable more robust training in challenging robotics settings.
\textbf{(2)}~We use tasks in legged locomotion, humanoid motion tracking, and both single-arm and bimanual manipulation to demonstrate settings where FPO++ succeeds: in learning policies from scratch, in sim-to-real transfer, and in \revision{finetuning} policies that are pretrained from demonstrations.
\textbf{(3)}~We ablate algorithmic choices,
showing that per-sample ratios consistently improve performance while the asymmetric trust region is important for entropy preservation when training from scratch. 
\textbf{(4)}~We analyze training dynamics and find that FPO++ learns correlated action distributions that diagonal Gaussian policies cannot represent, producing qualitatively different gaits from the same rewards and showing improved sample efficiency and robustness to base policy choice when \revision{finetuning}.

\section{Background and Related Work}

\textbf{Policy gradients and PPO.}
Policy gradient techniques are the dominant approach for continuous control with RL~\citep{lee2020learning,rudin2022learning,qi2023general,qi2025simple,hafner2025training}.
These algorithms are on-policy; rollouts in the form of observation, action, and reward tuples $(o_t, a_t, r_t)$ for each environment timestep $t$ are used to update a policy $\pi_\theta(a_t \mid o_t)$ to maximize expected return.
In the robotics community, the standard approach for achieving this is the clipped Proximal Policy Optimization (PPO)~\citep{schulman2017proximal} objective.
For an action with likelihood ratio $\rho_\theta$ and advantage estimate~\citep{schulman2015high} $\hat{A}_t$, this can be written as:
\begin{equation}
\psi_\text{PPO}\!\left(\rho_\theta, \hat{A}_t\right) = \min\!\left(\rho_\theta\hat{A}_t,\;
\text{clip}(\rho_\theta, 1 \pm \varepsilon^\text{clip})\hat{A}_t \right).
\label{eq:psi-ppo}
\end{equation}
The overall optimization problem is then
\begin{equation}
\max_\theta \;\;
\mathbb{E}_{\pi_{\theta_{\text{old}}}} 
\left[ \psi_\text{PPO}\!\left(\rho_\theta, \hat{A}_t\right) \right];\quad
\rho_\theta = \frac{\pi_\theta(a_t \mid o_t)}{\pi_{\theta_{\text{old}}}(a_t \mid o_t)}.
\label{eq:ppo}
\end{equation}
PPO is popular because it is simple to implement and provides strong empirical performance.
It also inherits advantages of general policy gradient algorithms:
unlike methods that rely on learned Q functions~\citep{lillicrap2015continuous} or transition models~\citep{hafner2019dream}, the PPO objective requires differentiability only from action likelihoods and not from rewards or environment dynamics. 

\textbf{Flow and diffusion policies.}
Flow and diffusion models~\citep{lipman2023flowmatchinggenerativemodeling,ho2020denoising}, which we consider equivalent in practice~\citep{gao2025diffusionmeetsflow}, are at the current frontier for supervised policy learning with continuous action spaces.
Systems relying on these models typically train robot manipulation policies from human demonstrations~\citep{chi2025diffusion,ankile2024imitation,robomimic2021}, and have been validated with language conditioning at larger scales~\citep{black2024pi0visionlanguageactionflowmodel,intelligence2025pi05,jiang2025streaming}. 
Similar approaches have also been adopted in whole-body humanoid control, where iterative generative policies are supervised by Gaussian experts~\citep{huang2025diffuse,liao2026beyondmimic}. 
All of these approaches need expert action labels.
Instead, we study online RL where supervision is only provided in the form of environment rewards.
This can be used to learn new behaviors or to \revision{finetune} pretrained policies.

\textbf{RL for flow and diffusion policies.}
Beyond imitation, recent works have also begun to integrate flow and diffusion policies into reinforcement learning settings.
Many of these works study offline RL from static datasets (Appendix~\ref{app:relwork_offline_rl}). %
We instead consider online RL using policy gradient-style training.
Existing work in this space is primarily focused on mechanisms for likelihood computation.
Similar to image diffusion RL techniques that formulate denoising as an MDP~\citep{black2023training,fan2023dpok}, DPPO~\citep{ren2024diffusion} and ReinFlow~\citep{zhang2025reinflow} optimize likelihoods computed from stochastic sampling noises. NCDPO~\citep{yang2025fine} optimizes likelihoods computed from both the initial noise and stochastic sampler noises, while backpropagating through unrolled denoising steps.
GenPO~\citep{ding2025genpo} also unrolls denoising steps, while incorporating an invertible architecture inspired by normalizing flows~\citep{dinh2016density}.
In contrast, we study a flow policy gradient~\citep{mcallister2025flow} approach that bypasses likelihoods entirely.
It does not rely on noise likelihoods from specific stochastic sampling trajectories, which (i) inflate the credit assignment horizon and (ii) are not equivalent to action likelihoods marginalized over initial noises and sampling trajectories.
It does not require specific network architectures or unrolling, which is expensive and risks vanishing or exploding gradients.
We describe how this is achieved in the next section.

\section{FPO++}

We introduce \ourmethod{}, an updated version of the FPO (Flow Policy Optimization) algorithm that succeeds in real-world robotics tasks.
We summarize FPO, then discuss \ourmethod{}.

\subsection{Preliminaries}

\textbf{Flow matching policy gradients.}
FPO~\citep{mcallister2025flow} aims to enable policy gradient-style training of policies parameterized as flow models~\citep{lipman2023flowmatchinggenerativemodeling}, without explicit likelihoods.
Although action likelihoods under flow policies can be computed by accounting for changes in volume, the direct or indirect divergence integration required for this~\citep{skreta2025superpositiondiffusionmodelsusing} is computationally prohibitive in RL settings.

FPO addresses this by proposing a surrogate for $\rho_\theta$,
\begin{equation}
\hat{\rho}_{\text{FPO}}(\theta) = 
\exp\!\Big( \hat{\mathcal{L}}_{\text{CFM},\theta_{\text{old}}}(a_t; o_t) 
- \hat{\mathcal{L}}_{\text{CFM},\theta}(a_t; o_t) \Big),
\label{eq:fpo-ratio}
\end{equation}
where $\hat{\mathcal{L}}_{\text{CFM},\theta}(a_t; o_t)$ is a Monte Carlo estimate of the conditional flow matching (CFM) loss.

This formulation enables PPO-style training of flow-based policies, which can express more complex distributions than the diagonal Gaussians that are most common in online reinforcement learning for robotics~\citep{rudin2022learning,qi2023general,videomimic}.
FPO mirrors PPO's clipped objective (\revision{Equation}~\ref{eq:ppo}), maximizing:
\begin{equation}
\max_\theta \;\;
\mathbb{E}_{\pi_{\theta_\text{old}}} 
\left[ \psi_\text{PPO}\!\left(\hat{\rho}_{\text{FPO}}(\theta), \hat{A}_t\right) \right].
\label{eq:fpo-obj}
\end{equation}
Intuitively, FPO's ratio approximation uses CFM loss differences to approximate action log-likelihood differences. %
The final objective (Equation~\ref{eq:fpo-obj}) then uses advantage estimates to shift probability flow toward higher-reward actions.

\textbf{Conditional flow matching loss.}
To estimate CFM losses, FPO first draws $N_\text{mc}$ noise $\epsilon_i \sim \mathcal{N}(0,I)$ and flow step $\tau_i \in [0, 1]$ pairs for each action $a_t$ and $i\in\{1 \dots N_\text{mc}\}$.
Noised actions are then computed using an interpolation schedule.
Linear interpolation is common in flow models:
\begin{align}
a_{t}^{\tau_i} &=  \tau_ia_t + (1 - \tau_i)\epsilon_i.
\end{align}
Squared errors are computed and averaged for the policy's velocity predictions $\hat{v}_\theta$,
\begin{align}
\hat{\mathcal{L}}_{\text{CFM},\theta}(a_t; o_t) &= 
\frac{1}{N_\text{mc}} \sum_{i}^{N_\text{mc}} \ell_{\theta}^{(i, t)} \\
\ell_\theta^{(i,t)} =
\ell_\theta(a_t, \tau_i, \epsilon_i; o_t) &= \left\| \hat{v}_\theta\!\left(a_t^{\tau_i}, \tau_i; o_t\right) - (a_t - \epsilon_i) \right\|^2_2.
\label{eq:cfm}
\end{align}
These losses can then be used in policy updates, which aim to increase or decrease CFM losses based on the sign of corresponding action advantages.
Without loss of generality, the network can also be trained to predict the clean action $a_t$, noise $\epsilon_i$, or a different combination of the pair~\citep{gao2025diffusionmeetsflow}.

\subsection{\ourmethod}

While the \revision{vanilla} FPO formulation succeeds in synthetic benchmarks~\citep{mcallister2025flow}, we found that it required refinements to achieve reliable performance in more difficult tasks.
\ourmethod{} proposes two changes to the FPO objective: (1)~per-sample ratios and (2)~an asymmetric trust region.

\textbf{Per-sample ratio.}
FPO estimates CFM losses by averaging over multiple $(\tau_i, \epsilon_i)$ samples for each action.
In the \revision{vanilla} FPO algorithm~\citep{mcallister2025flow}, this produces a single ratio per action:
\begin{equation}\label{eq:per_action_ratio}
\hat{\rho}_\text{FPO}(\theta) 
= \exp\!\left( \tfrac{1}{N_\text{mc}}\sum_{i=1}^{N_\text{mc}} 
  \bigg(\ell_{\theta_{\text{old}}}^{(i,t)}
     - \ell_\theta^{(i,t)} \bigg)\right).
\end{equation}
An important characteristic of this formulation is that ratios are clipped \textit{after} averaging across samples.
For a given action, this means that either all or no samples are clipped.
In \ourmethod{}, we instead calculate a separate ratio for each sample $i$,
\begin{equation}\label{eq:per_sample_ratio}
\hat{\rho}_\text{FPO++}^{(i)}(\theta) 
=  \exp\!\left(\ell_{\theta_{\text{old}}}^{(i,t)} 
              - \ell_\theta^{(i,t)} \right).
\end{equation}
The same advantage $\hat{A}_t$ is shared across samples.
Equations~\ref{eq:per_action_ratio} and \ref{eq:per_sample_ratio} produce identical gradients for on-policy data, where all ratios evaluate to 1.
When taking multiple gradient steps, however, the per-sample ratio provides a finer-grained trust region than the original per-action formulation.
It allows each $(\tau_i,\epsilon_i)$ pair to be clipped independently. 

\textbf{Asymmetric trust region (ASPO).}
We found that the stability of FPO when training policies from scratch can be improved by adjusting its trust region implementation.
For solving these tasks in FPO++, we introduce an asymmetric trust region that we refer to as \textit{Asymmetric SPO (ASPO)}.
We use PPO clipping (Equation~\ref{eq:psi-ppo}) for positive-advantage actions where gradients push to decrease CFM losses; for negative-advantage actions where gradients push to increase the CFM loss, we adopt the more constrained Simple Policy Optimization (SPO) objective proposed by~\cite{xie2024simple}:
\begin{equation}
\psi_\text{SPO}\!\left(\rho_\theta, \hat{A}_t\right) =
\rho_\theta\,\hat{A}_t
  \;-\; \frac{|\hat{A}_t|}{2\,\varepsilon^{\mathrm{clip}}}\,\big(\rho_\theta-1\big)^2.
\label{eq:psi-spo}
\end{equation}
Instead of zeroing out gradients for samples with ratios that surpass the trust region, the SPO objective provides a gradient signal that pulls ratios back.

Applying SPO to negative advantages disincentivizes large CFM loss increases during FPO++ updates.
When interpreting the CFM loss as a variational bound~\citep{mcallister2025flow,kingma2023understandingdiffusionobjectiveselbo},
this penalizes (i)~aggressive decreases in action likelihoods and (ii)~aggressive increases in the KL divergence between the sampled and learned denoising posteriors.
The first property is attractive for preserving entropy~\citep{yu2025dapo}, while the second property stabilizes the variational gap.

\subsection{\ourmethod{} Objective}
\label{sec:final_objective}

We now summarize the \ourmethod{} objective by combining the modifications described above. 
For each action $a_t$ with advantage $\hat{A}_t$, we draw $N_\text{mc}$ Monte Carlo pairs $(\tau_i,\epsilon_i)$. 
The ASPO trust region combines Equations~\ref{eq:psi-ppo} and \ref{eq:psi-spo}, defined piecewise based on the sign of the advantage:
\begin{equation}
\psi_\text{ASPO}\!\left(\rho_\theta, \hat{A}_t\right) =
\begin{cases}
\psi_\text{PPO}\!\big(\rho_\theta, \hat{A}_t\big),
& \hat{A}_t \geq 0, \\[2pt]
\psi_\text{SPO}\!\big(\rho_\theta, \hat{A}_t\big), 
& \hat{A}_t < 0.
\end{cases}
\end{equation}
The \ourmethod{} objective is then
\begin{equation}
\max_\theta \;\; 
\mathbb{E}_{\pi_{\theta_{\text{old}}}} 
\left[ \sum_{i=1}^{N_\text{mc}}
\; \psi_\text{ASPO}\!\left(\hat{\rho}^{(i)}_{\text{FPO++}}(\theta), \hat{A}_t \right)
\right].
\end{equation}

\section{Experiments}

This section summarizes our main findings and discussion, with additional supporting figures and analysis provided in the appendix and referenced inline where relevant.

\subsection{Locomotion Benchmarks}

\begin{figure*}[t!]
    \centering
    \includegraphics[width=0.9\textwidth]{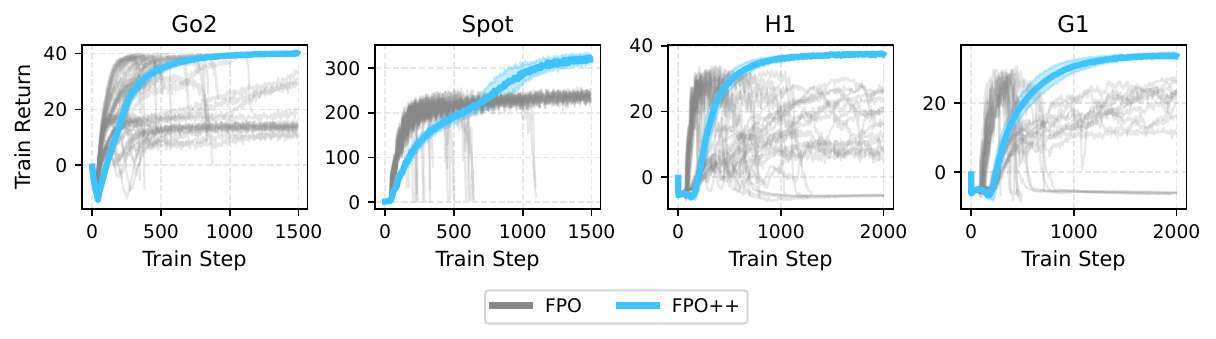}
    \caption{
        \textbf{Improved stability in IsaacLab locomotion environments.}
        We compare episode returns from FPO++ training with FPO; FPO++ curves show the mean over 5 random seeds.
        For thoroughness and to ensure we are not comparing against a bad set of hyperparameters for the FPO baseline, we include FPO runs for many different hyperparameter choices: all combinations of learning rate $\in \{10^{-5}, 10^{-4}, 3\times 10^{-4}\}$, clip parameter $\in \{0.04,0.05,0.06\}$, and Monte Carlo samples $\in \{8, 16, 32\}$.
        FPO++ addresses stability problems that we were unable to solve by tuning FPO hyperparameters.
    }
    \label{fig:locomotion_training_curves}
\end{figure*}

\begin{figure*}[t]
    \centering
    \subfloat[T1 locomotion\label{fig:t1_locomotion}]{%
        \makebox[0.25\linewidth][c]{\includegraphics[height=4.25cm]{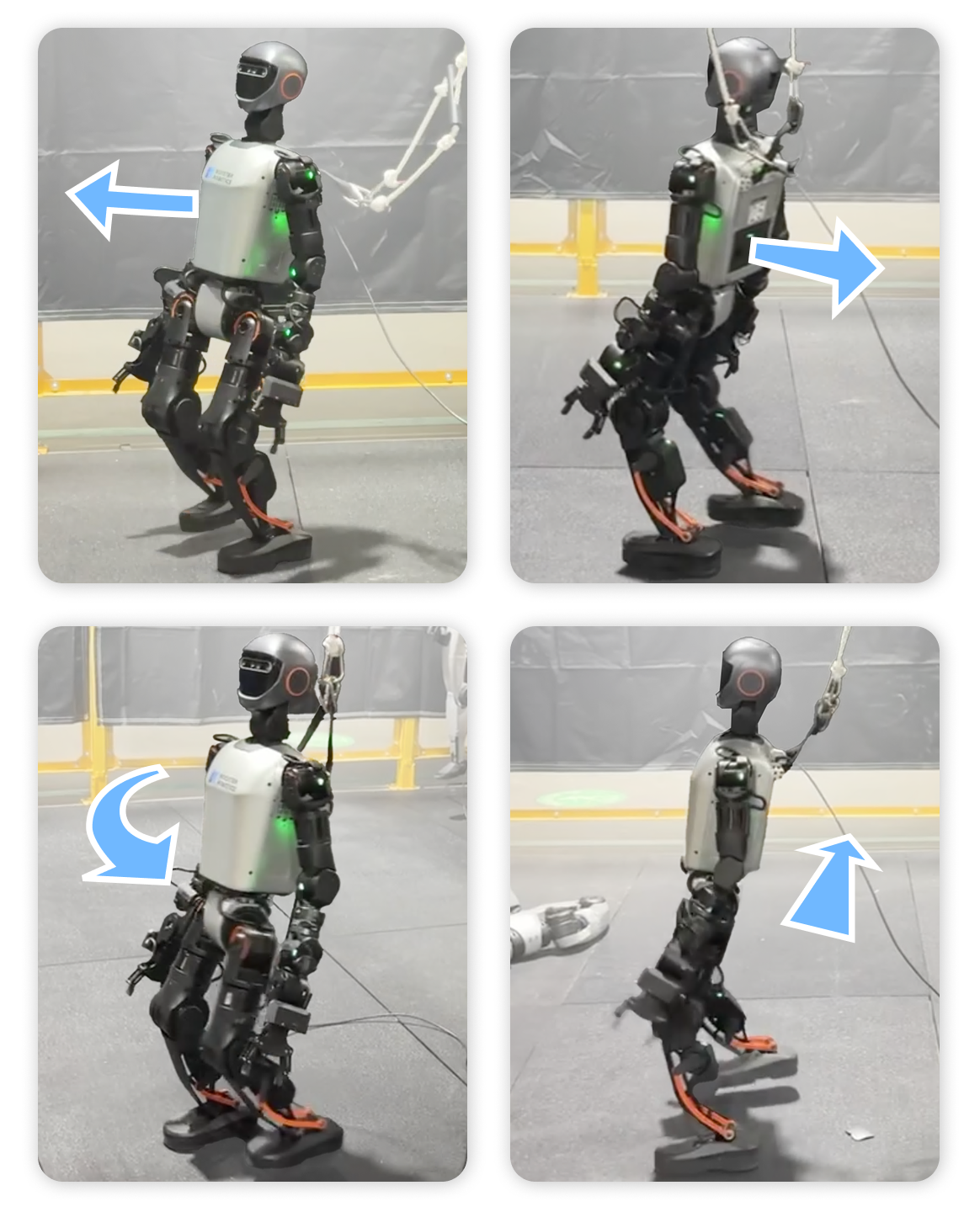}}%
    }%
    \subfloat[G1 motion tracking\label{fig:g1_motion_tracking}]{%
        \makebox[0.5\linewidth][c]{
        \includegraphics[height=4.25cm]{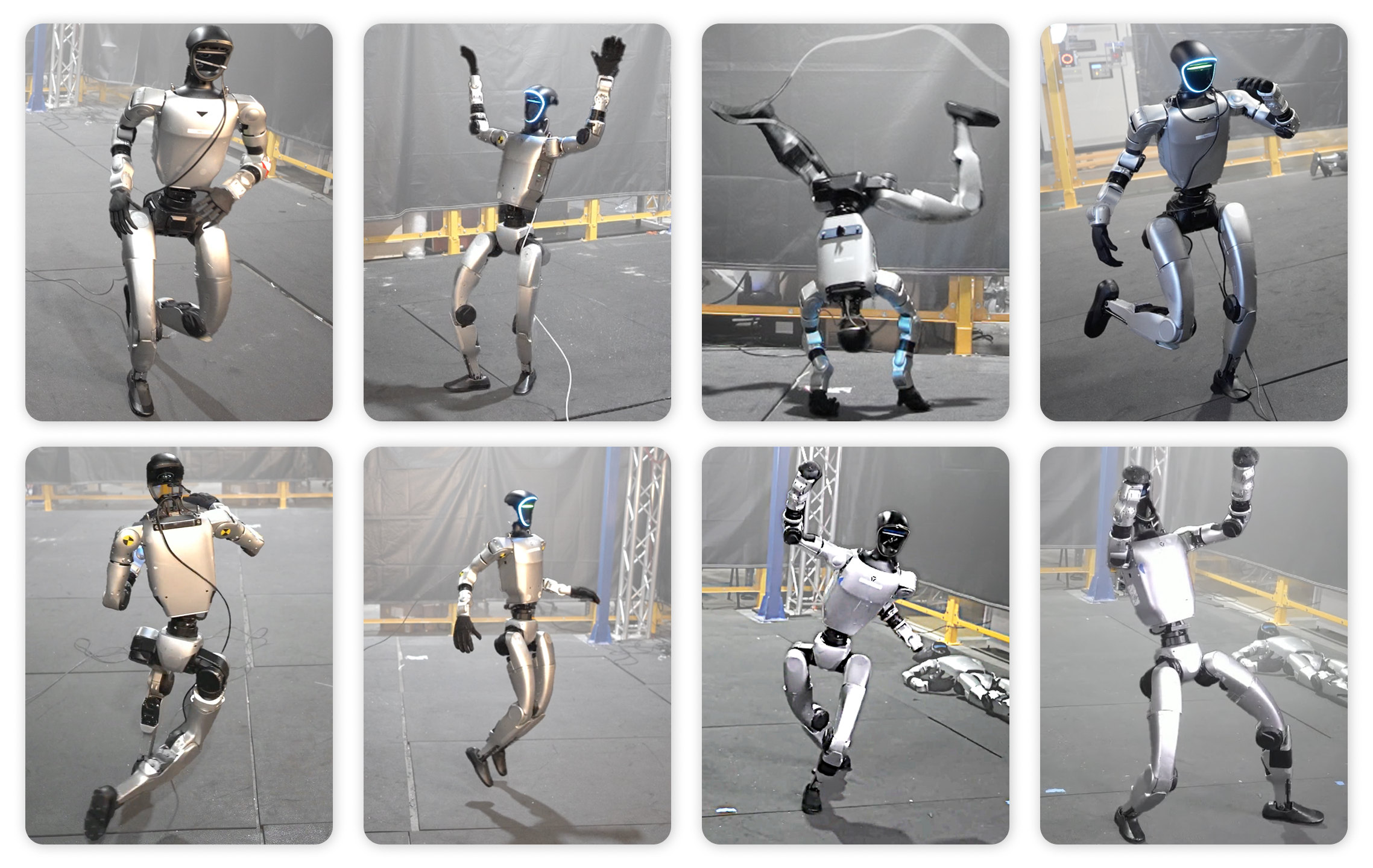}%
        }
    }%
    \subfloat[G1 robustness test\label{fig:G1_robust_test}]{%
        \makebox[0.25\linewidth][c]{
        \includegraphics[height=4.25cm]{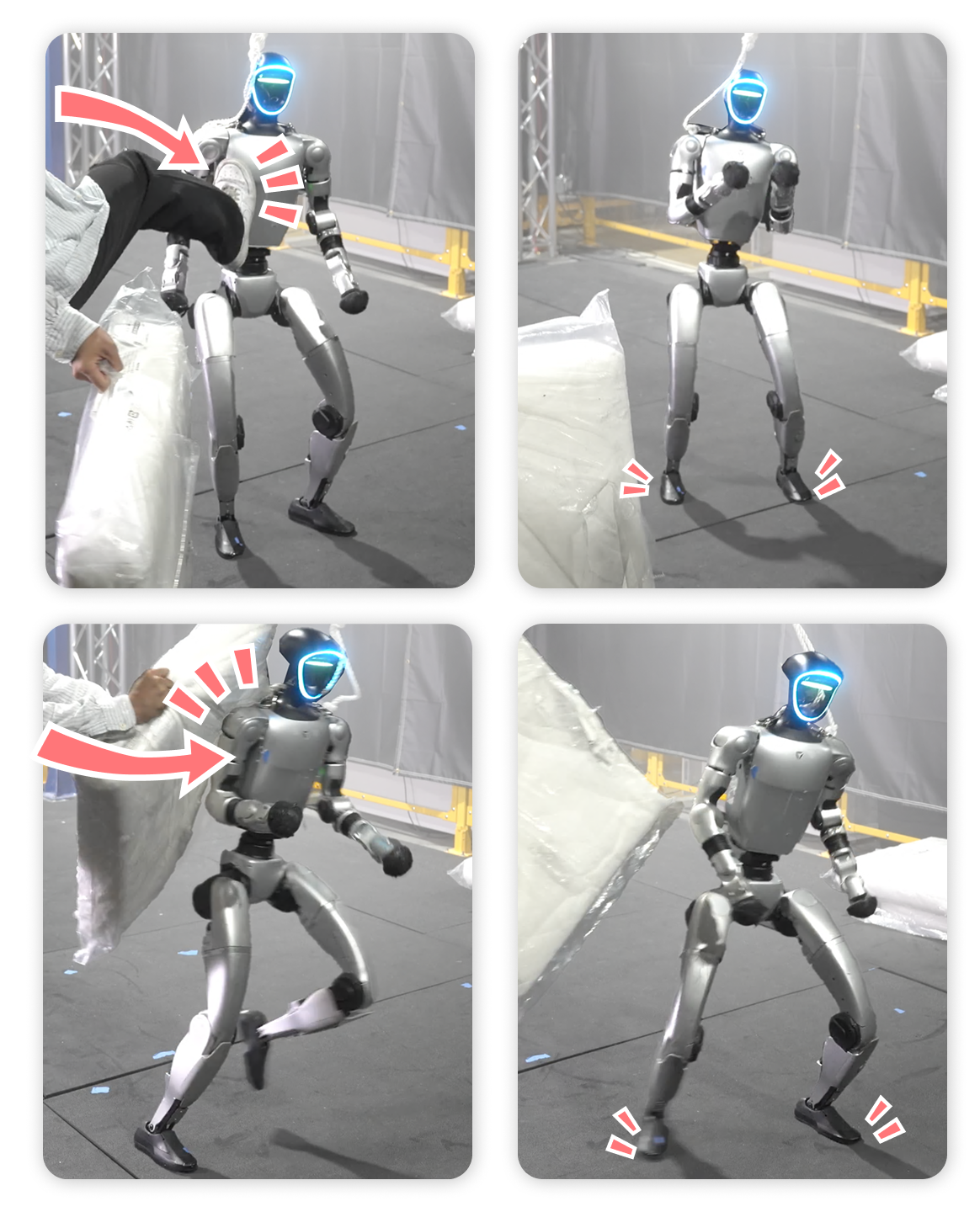}%
        }
    }%
    \caption{
        \textbf{Sim-to-real transfer.}
        We deploy flow policies for locomotion to a Booster T1 and motion tracking to a Unitree G1.
        Policies are directly deployed to real robots with reduced sampling step counts, demonstrating stable gaits, tracking for long sequences, and robustness to external forces. The arrows in T1 locomotion (Figure~\ref{fig:t1_locomotion}) indicate velocity commands, while the arrows in the robustness tests (Figure~\ref{fig:G1_robust_test}) highlight external forces.
    }
    \label{fig:sim2real}
\end{figure*}

\textbf{Experiment setup.}
In our first set of experiments, we train policies using the standard IsaacLab~\citep{mittal2023orbit} velocity-conditioned robot locomotion environments.
We include results on four different robots: two quadrupeds (Unitree Go2, Boston Dynamics Spot) and two humanoids (Unitree H1 and G1).
Policies in these environments take proprioceptive state, linear velocity target, and angular velocity target as input, and are rewarded for matching the given velocity targets.

\textbf{FPO++ dramatically improves stability.}
Figure~\ref{fig:locomotion_training_curves} reports FPO and FPO++ training curves on each of the four robots.
We found that \revision{vanilla} FPO was more prone to local minima and catastrophic failures in these environments than the DeepMind Control Suite~\citep{tassa2018deepmind,zakka2025mujoco} or PHC~\citep{luo2023perpetual} tasks evaluated by prior work~\citep{mcallister2025flow}.
We attribute this to a combination of factors: high-dimensional action spaces, realistic joint position and torque limits, and coarser reward functions; instability persisted even with hyperparameter tuning, gradient clipping, and observation normalization.
In contrast, FPO++ achieves and then maintains high episode returns.

\subsection{Humanoid Sim-to-real}

\textbf{Experiment setup.}
Next, we consider sim-to-real transfer of flow policies using two humanoid robots: the Booster T1 and Unitree G1.
We modify HumanoidVerse~\citep{HumanoidVerseGithub,makoviychuk2021isaac} to train and deploy flow policies for velocity-conditioned T1 locomotion, and adapt the IsaacLab-based BeyondMimic~\citep{liao2026beyondmimic} codebase to train and deploy flow policies for whole-body G1 motion tracking.
Locomotion policies are conditioned on and trained to match linear and angular velocity targets; motion tracking policies are conditioned on and trained to match retargeted reference motions from the LAFAN dataset~\citep{harvey2020robust}.
We use six reference motions: \textit{dance1\_subject2},\ \textit{dance1\_subject1},\ \textit{walk1\_subject1},\ \textit{run1\_subject2},\ \textit{fight1\_subject2},\ and\ \textit{jumps1\_subject1}, which cover a wide variety of dynamic whole-body control challenges and last around 2.5 minutes each on average.
We train one policy per motion and deploy each on the real robot, following the RL-based motion tracking protocol of BeyondMimic~\citep{liao2026beyondmimic}.

\textbf{FPO++ policies succeed on physical robots.}
FPO++ learns robust gaits for T1 locomotion and reliably tracks dynamic motion sequences on the G1, deployed as shown in Figure~\ref{fig:sim2real}.
We consider this a significant result: to our knowledge, it is the first demonstration of humanoid sim-to-real using either (i) a flow policy trained without expert distillation or (ii) a policy gradient technique without explicit likelihoods.

\begin{figure*}[t!]
    \centering
    \subfloat[Per-sample ratios improve locomotion policies.\label{fig:ablation_ratio}]{%
        \includegraphics[width=0.8\textwidth]{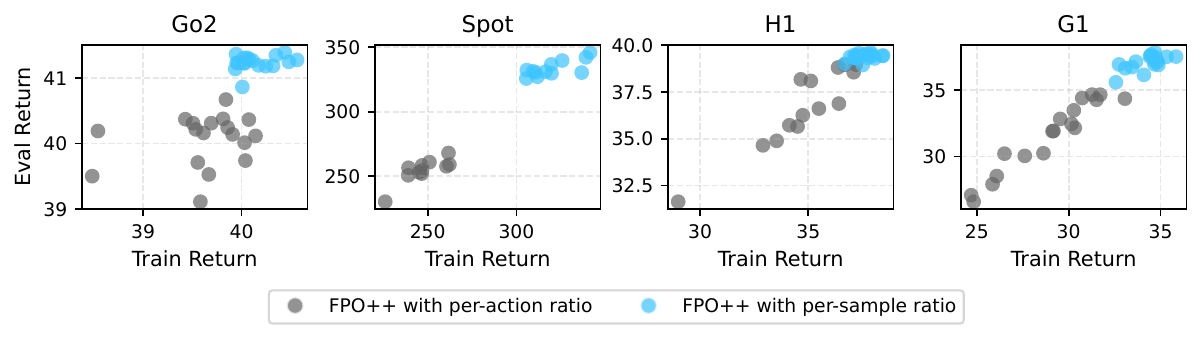}%
    }

    \subfloat[ASPO produces higher, more consistent returns.\label{fig:ablation_trust_region}]{%
        \includegraphics[width=0.8\textwidth]{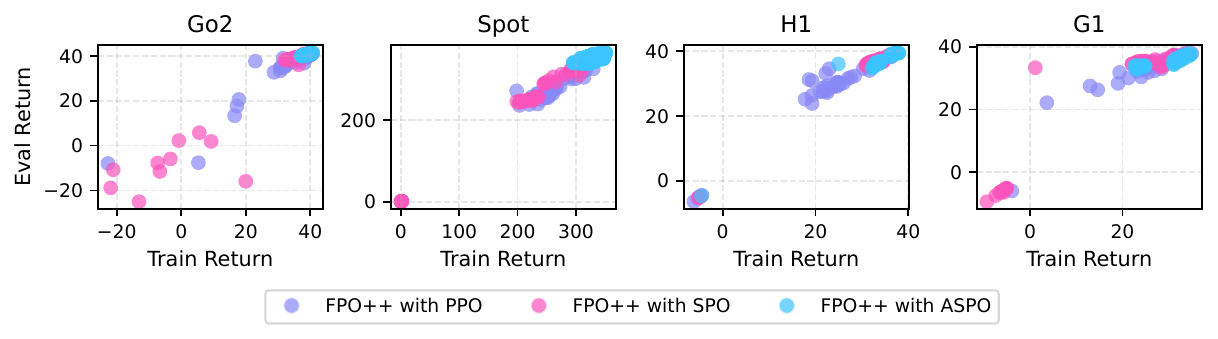}%
    }
    \caption{
        \textbf{FPO++ ablations on locomotion.}
        We plot final training and evaluation returns across many runs; each point is a training run with clipping parameter $\in \{0.04, 0.05, 0.06\}$ and random seed $\in \{0,1,2,3,4\}$.
        \textbf{(a)}~compares FPO++ with per-sample ratios (Equation~\ref{eq:per_sample_ratio}) against per-action ratios from prior work (Equation~\ref{eq:per_action_ratio}); per-sample ratios produce higher and more consistent returns.
        \textbf{(b)}~compares trust region implementations: standard PPO clipping, SPO~\citep{xie2024simple}, and our asymmetric ASPO objective.
    }
    \label{fig:ablations}
\end{figure*}

\begin{figure*}[t!]
    \centering
    \includegraphics[width=0.8\textwidth]{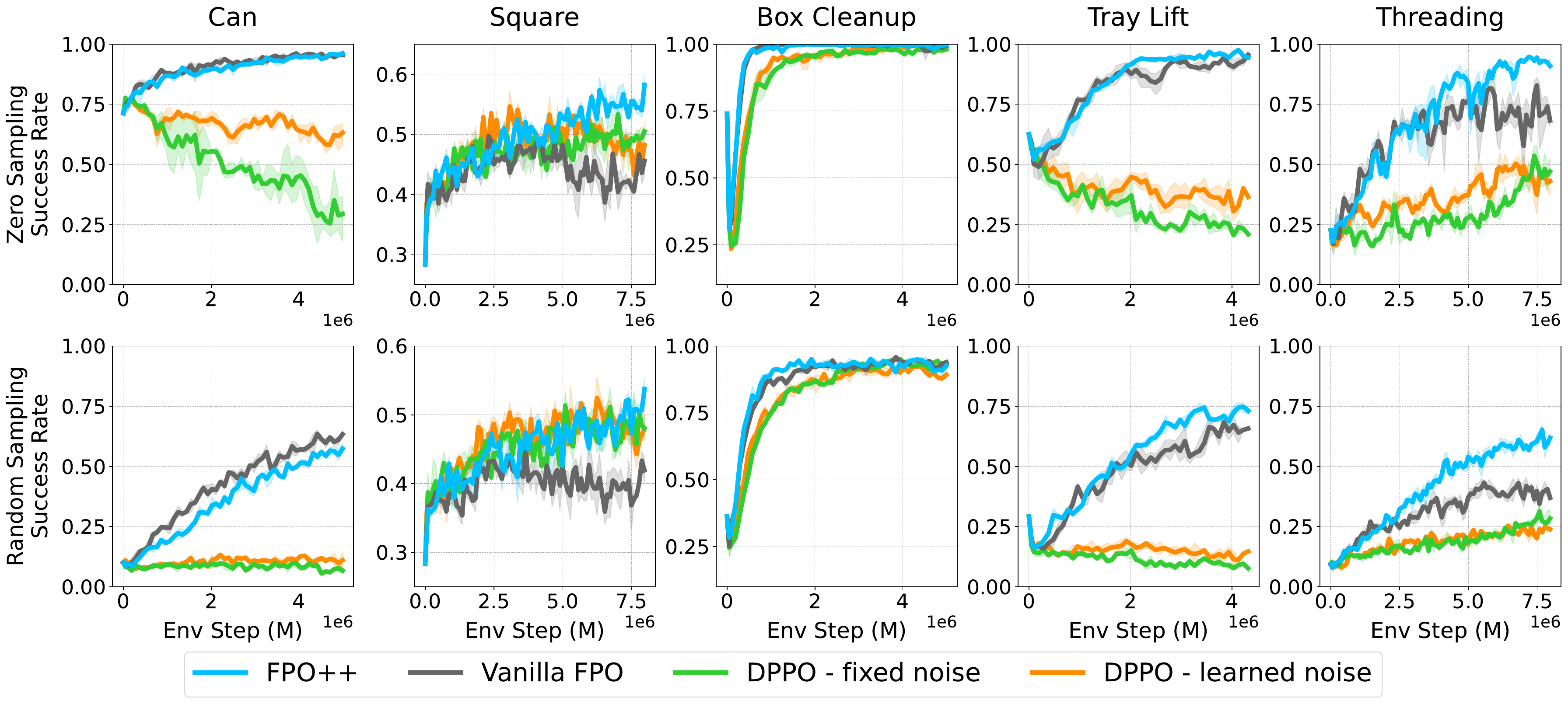}
    \caption{
    \textbf{Manipulation \revision{finetuning} results.}
    We compare the evaluation success rates for FPO++, FPO, and our two DPPO implementations.
    We show two rows: the first row contains policy success rates with zero-sampling ($\epsilon = \vec{0}$), while the second row contains policy success rates using standard random sampling ($\epsilon \sim \mathcal{N}(0,I)$). For each task, all algorithms are initialized from the same base policy, which is an image-based flow matching policy trained to predict action chunks.
}
\label{fig:manipulation_comparison}
\end{figure*}

\subsection{Manipulation \revision{Finetuning}}
\label{sec:manipulation_finetuning}

\textbf{Experiment setup.}
One application not studied in prior flow policy gradient work~\citep{mcallister2025flow} is reward-based \revision{finetuning} for policies that are first trained from demonstrations.
The success of flow policies in imitation learning~\citep{black2024pi0visionlanguageactionflowmodel} makes RL of flow policies uniquely important in this setting.
To validate FPO++ for \revision{finetuning}, we begin by pretraining image-based manipulation policies for tasks from RoboMimic~\citep{robomimic2021} and DexMimicGen~\citep{jiang2025dexmimicgen}.
We follow the data processing of ResFiT~\citep{ankile2025residual} for five tasks, which cover diverse embodiments and control regimes (Figure~\ref{fig:manipulation_tasks}).
All policies use a 3-layer MLP backbone with a ViT~\citep{dosovitskiy2020vit} encoder for image observations.
We then \revision{finetune} flow policies using FPO++, FPO, and two baselines adapted from the DPPO~\citep{ren2024diffusion} implementation in~\citep{mcallister2025flow}: one with a fixed noise scale for exploration and one with a predicted noise inspired by ReinFlow~\citep{zhang2025reinflow} (shown as \emph{DPPO -- fixed noise} and \emph{DPPO -- learned noise} in Figure~\ref{fig:manipulation_comparison}).
Implementation and baseline details are in Appendix~\ref{app:manipulation_more_baselines}; notably, we disable ASPO for manipulation experiments (Section~\ref{sec:exp_ablations}).

\textbf{Flow policy gradients succeed in \revision{finetuning}.}
We plot policy success rates using each \revision{finetuning} algorithm in Figure~\ref{fig:manipulation_comparison}.
We find that FPO++ consistently achieves high success rates, converging more rapidly than baselines.
Vanilla FPO also performs well on simpler manipulation tasks, suggesting that behavior cloning initialization can provide sufficient regularization for mitigating the instabilities we observed when training from scratch.
Both DPPO variants underperform FPO-based methods, which we attribute to the longer effective MDP horizon introduced by treating diffusion steps as decision points~\citep{yang2025fine}.
These results are notable given the emphasis on likelihood computation in prior flow and diffusion \revision{finetuning} methods (e.g., two-layer MDPs~\citep{ren2024diffusion} and learned noise predictors~\citep{zhang2025reinflow}); FPO++ suggests the explicit density estimates motivating these changes are not necessary for effective flow RL training.

\textbf{FPO++ is more robust to base-policy quality.}
We found that our DPPO runs often failed when base policies had low \revision{random sampling} success rates, like the Can example in Figure~\ref{fig:manipulation_comparison}, which starts at $\sim10\%$ success rate.
The same task succeeds when initialized with higher-quality base policies; examples are provided in Appendix~\ref{app:can_success_rate_analysis}.

\subsection{Training Ablations}
\label{sec:exp_ablations}

The experiments above demonstrate stable flow policy learning across locomotion, sim-to-real, and manipulation tasks.
To better understand what makes this possible, we ablate FPO++'s per-sample ratio and ASPO trust region.

\textbf{FPO++ changes are critical for all locomotion embodiments.}
Figure~\ref{fig:ablation_ratio} shows final training and evaluation returns for locomotion environments when we ablate the per-sample ratio; Figure~\ref{fig:ablation_trust_region} shows returns when we ablate the ASPO trust region.
We include multiple hyperparameter configurations for thoroughness: each point is a run trained using a unique clipping parameter and random seed combination.
We find that FPO++ achieves consistent policy performance across hyperparameters and random seeds.
When either the per-sample ratio or ASPO trust region is disabled, average returns drop significantly across robot embodiments.
The variance of returns also increases, suggesting that the ratio and ASPO trust region combination stabilizes learning.

\textbf{ASPO successfully preserves entropy.}
We compare flow field visualizations for policies trained using PPO and ASPO trust regions in Appendix~\ref{app:full_flow_field} (Figure~\ref{fig:entropy_small}).
Policies trained using ASPO successfully solve tasks without entropy collapse, explaining the improved robustness in Figure~\ref{fig:ablation_trust_region}.

\textbf{FPO++ reduces empirical gradient variance.}
A motivation for FPO++'s changes is reduced gradient variance, which we verify using a cosine similarity metric inspired by~\cite{ilyas2018closer} (Appendix~\ref{app:measuring_variance}).

\revision{\textbf{ASPO's benefit depends on the reward regime.}
ASPO improves locomotion but underperforms PPO in manipulation \revision{finetuning} (Appendix~\ref{app:manipulation_ablation}). To test whether this stems from pretraining, we compare both trust regions in FPO++ on RoboMimic Lift with randomly initialized action networks (Appendix~\ref{app:lift_reward_regime}, Figure~\ref{fig:lift_reward_regime}). With sparse rewards, ASPO barely learns, reaching 0.07 final success versus PPO's 0.73. This suggests that limiting likelihood decreases for negative-advantage actions keeps the policy near its random initialization and hinders bootstrapping from rare successes. With dense rewards, ASPO initially learns but later loses success despite relatively high shaped return (0.28 final success versus PPO's 0.81). Rollouts reveal grasp-and-hover behavior that earns shaping reward without completing the lift. These results suggest that ASPO's effectiveness depends on the reward regime, not just policy initialization. We recommend ASPO for dense rewards closely aligned with task completion, as in locomotion and motion tracking, PPO for sparse rewards, and caution with ASPO when shaping rewards can be exploited.}

\subsection{Comparison with Gaussian PPO}

While some FPO++ use cases (e.g., the \revision{finetuning} in Section~\ref{sec:manipulation_finetuning}) require flow policies, other tasks are designed for Gaussian PPO---a well-understood baseline for validating FPO++ training, analyzing dynamics, and assessing sample efficiency and expressiveness.
We investigate these characteristics by first comparing FPO++ against IsaacLab's default Gaussian PPO implementation~\citep{schwarke2025rslrl} across locomotion tasks with varying amounts of parallelization (Figure~\ref{fig:ppo_compare_numenvs}).
Details and hyperparameter tuning information can be found in Appendix~\ref{app:experiment_details_locomotion}.

\textbf{FPO++ locomotion policies show sample efficiency advantages.}
We observe in Figure~\ref{fig:ppo_compare_numenvs} that FPO++ locomotion policies for each environment count almost always converge to higher returns than the Gaussian PPO configurations we compare against, with reduced variance between random seeds.
In particular, we see improved robustness to very small batch sizes in Go2, H1, and G1, and better exploitation of increased parallelism for Spot.

\begin{figure}[t]
    \centering
    \includegraphics[width=0.65\linewidth]{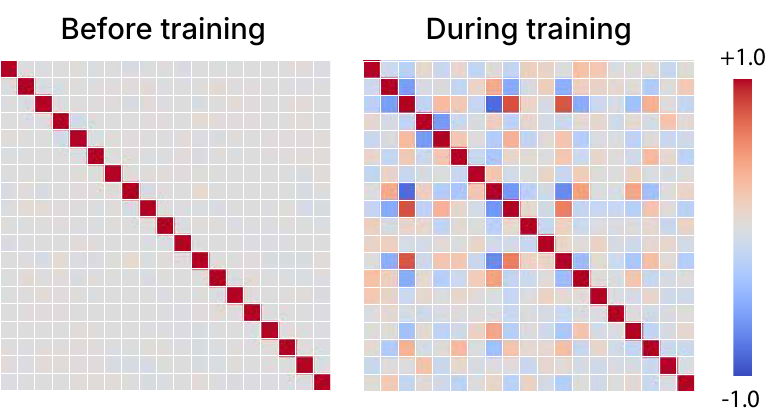}
    \caption{
        \textbf{Correlations before and during training.}
        We sample 10,000 actions conditioned on the same observation, and visualize the correlations between action dimensions before and during FPO++ training for an H1 locomotion policy.
    }
    \label{fig:correlation}
\end{figure}

\textbf{FPO++ policies explore with more expressive action distributions.}
Consistent with~\cite{seo2025fasttd3simplefastcapable}, we found that the same rewards often produced different gaits with different algorithms.
FPO++ policies trained with default Spot locomotion rewards, for example, produced more consistent ``trot'' gaits than Gaussian PPO policies, which tended toward more symmetric ``pronk'' gaits (Figure~\ref{fig:ppo_fpo_spot_gait}).
To examine dependencies between action dimensions, Figure~\ref{fig:correlation} visualizes correlations for an H1 locomotion policy.
Off-diagonal correlations are near zero before training but develop structured positive and negative patterns during training, including negative correlations between the left and right hip joints.
Because actions are sampled at a fixed observation, these patterns reflect dependencies within the policy's conditional action distribution rather than changes in robot state along a rollout.
They show that FPO++ learns to vary joint commands together or in opposition during stochastic exploration, unlike diagonal-Gaussian PPO policies whose action dimensions are conditionally independent.

\textbf{Limitations and future work.}
FPO++ provides a single algorithm and policy representation that succeeds across embodiments and tasks, for both training from scratch and \revision{finetuning}, but it is not fully turnkey and some challenges remain.
\revision{The benefit of ASPO depends on the reward regime rather than initialization alone (Section~\ref{sec:exp_ablations}); choosing the right configuration still requires task-specific knowledge.}
\revision{Gaussian PPO with entropy regularization and an adaptive learning rate also achieves higher motion-tracking returns (Appendix~\ref{app:gaussian_motion_tracking}).}
FPO++ also takes more wall-clock time than Gaussian PPO, limiting its appeal where Gaussian policies already suffice: for G1 locomotion on an L40S GPU, our Gaussian PPO baseline reaches an evaluation return of 25 in 19 minutes, versus 23 minutes for FPO++, and motion tracking can take up to \revision{$2\times$} longer than a tuned Gaussian PPO baseline.
Reducing this gap, for example through few-step distillation~\citep{salimans2022progressive,liu2022flow} for more efficient training and inference, is promising future work.
Other directions include applications where Gaussian policies do not apply, such as tasks requiring more expressive exploration or diffusion-based sequence modeling~\citep{huang2025diffuse,chen2024diffusion}.

\section{Conclusion}

We presented a study of robot control with flow matching policy gradients.
The proposed FPO++ algorithm is more stable than prior flow policy gradient implementations and succeeds on practical locomotion, motion tracking, and manipulation finetuning tasks.
Our analysis highlights the effect of per-sample ratios and an asymmetric trust region, and shows how FPO++ exploits the flow representation: learning correlated action distributions and distinct gaits, with improved sample efficiency and \revision{finetuning} robustness over baselines.

Beyond validating a specific algorithm on specific tasks, FPO++ aims to serve as an existence proof that we are excited about for several reasons. FPO++ validates several possibilities, including policy gradient-style training of flow policies for real-world continuous control and sim-to-real transfer of these policies after training using only RL. Importantly, these results challenge the common assumption that explicit likelihoods are needed for policy gradient methods in robot control. By demonstrating that this constraint can be bypassed, FPO++ suggests new directions for expanding the design space of RL algorithms for future robot learning systems.

\acknowledgments{\revision{We thank David McAllister, Arthur Allshire, and Chungmin Kim for helpful discussions.}}

\bibliography{references}

\clearpage
\appendix
\renewcommand{\thesection}{\Alph{section}}

\renewcommand{\thesubsection}{\Alph{section}.\arabic{subsection}}

\renewcommand{\thefigure}{A.\arabic{figure}}
\renewcommand{\thetable}{A.\arabic{table}}
\renewcommand{\theequation}{A.\arabic{equation}}
\setcounter{figure}{0}
\setcounter{table}{0}
\setcounter{equation}{0}

\vspace*{1.0cm}
\begin{center}
  {\Large \bfseries Appendix of ``Flow Policy Gradients for Robot Control''\par}
  \vspace{0.6cm}
\end{center}

\begin{center}
\begin{minipage}{\linewidth}
    \centering
    \includegraphics[width=.9\linewidth]{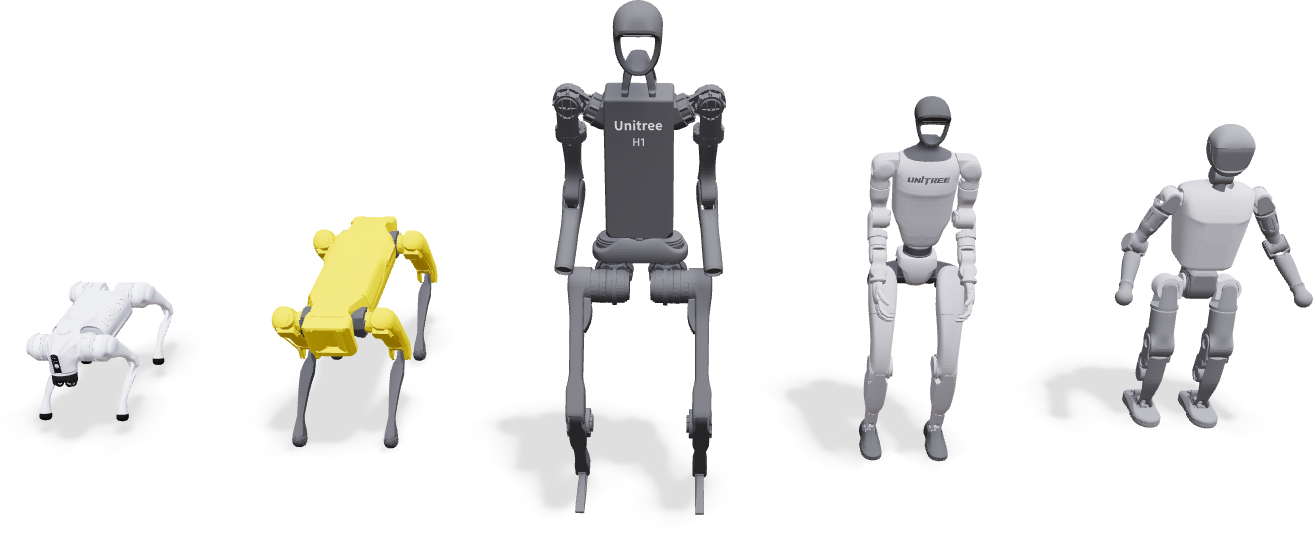}
    \captionof{figure}{
        \textbf{Legged robots used for experiments.}
        We train policies for the Go2, Spot, H1, and G1 robots in simulation.
        We deploy policies to physical G1 and Booster T1 robots.
    }
    \label{fig:legged_robots}
\end{minipage}
\end{center}

\begin{figure*}[h!]
    \centering
    \subfloat[Can\label{fig:can}]{%
        \includegraphics[height=2.5cm]{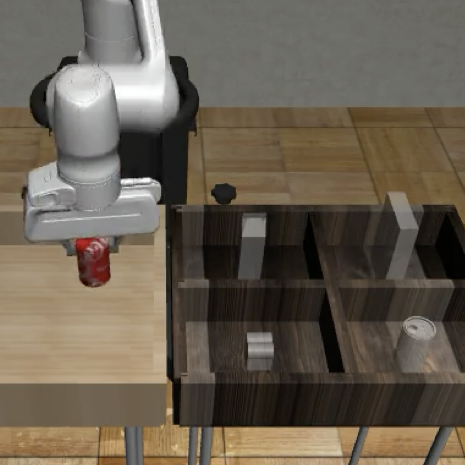}%
    }
    \hfil
    \subfloat[Square\label{fig:square}]{%
        \includegraphics[height=2.5cm]{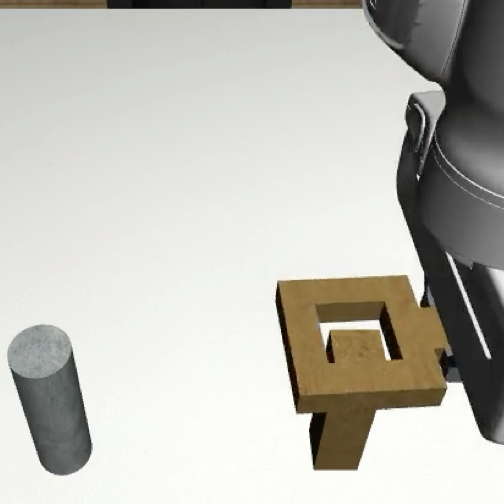}%
    }
    \hfil
    \subfloat[Box Cleanup\label{fig:boxcleanup}]{%
        \includegraphics[height=2.5cm]{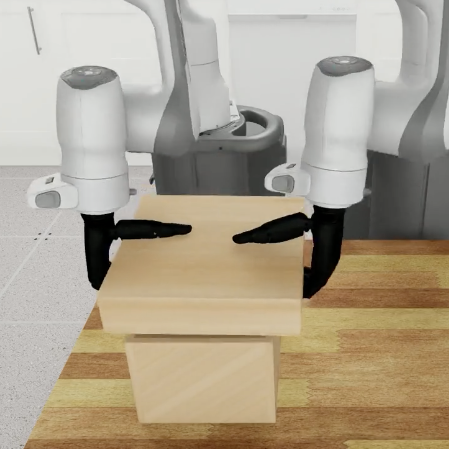}%
    }
    \hfil
    \subfloat[Tray Lift\label{fig:traylift}]{%
        \includegraphics[height=2.5cm]{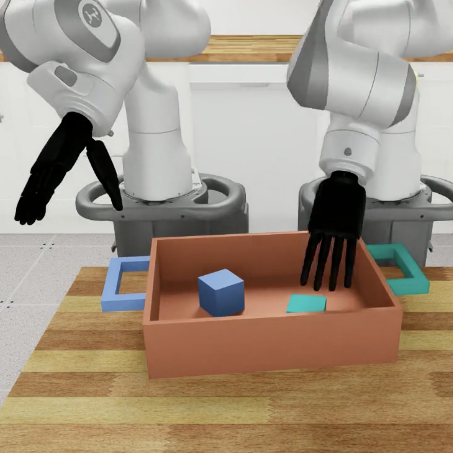}%
    }
    \hfil
    \subfloat[Threading\label{fig:threading}]{%
        \includegraphics[height=2.5cm]{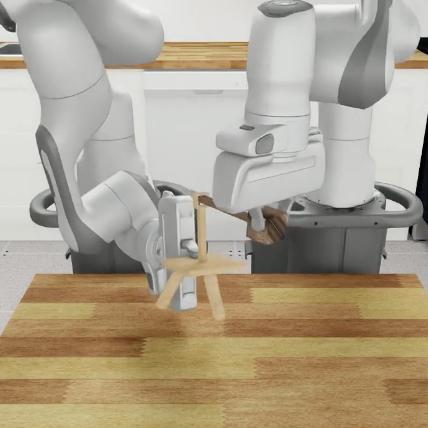}%
    }
    \vspace{0.3em}
    
    \makebox[0.36\linewidth][c]{RoboMimic}
    \makebox[0.54\linewidth][c]{DexMimicGen}
    \vspace{0.75em}
    \caption{
        \textbf{Manipulation tasks.}
        We compare FPO++ to baseline methods on manipulation tasks from RoboMimic~\citep{robomimic2021} and DexMimicGen~\citep{jiang2025dexmimicgen}.
        We choose tasks that cover diverse embodiments and control regimes: from single-arm to bimanual manipulation, parallel-jaw grippers and dexterous hands, and both short- and long-horizon tasks.
    }
    \label{fig:manipulation_tasks}
\end{figure*}

\section{Further related work}
\label{app:relwork_offline_rl}

The main body of our paper discusses online RL, where policies are updated from their own experience.
A related body of work has explored using flow and diffusion-based policies in offline RL. One common strategy is advantage weighted regression (AWR)~\citep{peng2019advantage,kang2024efficient,ding2024diffusion,zhang2025energy}. Another line of work optimizes a Q-learning objective jointly with a generative-model loss~\citep{wang2022diffusion,lu2023contrastive,he2023diffcps,ding2023consistency,zhang2024entropy,ada2024diffusion}, enabling value-based training while regularizing the policy through diffusion or flow matching. Maximum-entropy approaches such as DIME~\citep{celik2025dime} extend this direction by integrating diffusion policies with entropy-regularized RL, further improving robustness and sample quality in offline settings.

A key challenge in training diffusion or flow policies with Q-learning is that backpropagation through the multi-step denoising process is numerically unstable. FQL~\citep{park2025flow} addresses this by training a one-step flow policy without backpropagation through time (BPTT). Q-score matching~\citep{psenka2023learning} links the score of the diffusion policy to the action gradient of the Q-function.
Finally, QAM~\citep{li2026q} leverages adjoint matching~\citep{havens2025adjoint} to transform the critic's action gradient into a step-wise objective for the policy.

\section{Experiment Details}

In this section, we detail the experimental setup for the IsaacLab velocity-conditioned locomotion benchmarks, motion tracking benchmarks, and manipulation finetuning benchmarks used throughout the paper. The robots we used in the paper, including quadrupeds, humanoids, and manipulators, are illustrated in Figures~\ref{fig:legged_robots} and~\ref{fig:manipulation_tasks}.

\subsection{Locomotion Benchmarking}
\label{app:experiment_details_locomotion}

\subsubsection{Hyperparameters}

All policies for the locomotion benchmarking experiments utilize 3-layer Multi-Layer Perceptrons (MLPs). Specifically, the actor network employs 256 hidden units, and the critic network employs 768 hidden units per layer.
Following the IsaacLab defaults, we train with 4096 parallel environments and take 24 environment steps between policy updates, and we use 64 Euler steps for all rollouts.
Lower step counts (8, 16) reduced training stability without significantly impacting runtime: policy inference is fast, requires no gradients (even for training rollouts), and is not a training bottleneck.
Additional hyperparameters can be found in Table~\ref{tab:flow_rsl_rl_hyperparameters_locomotion}.

Quadruped policies are trained for 1500 steps, and humanoid policies are trained for 2000 steps.
For the Gaussian PPO results, we adopt the default hyperparameters provided by the \texttt{rsl\_rl} library, with additional sweeps performed over the clipping parameters in the set $\{0.1, 0.15, 0.2, 0.25\}$.

\subsubsection{Environments and Rewards}

All locomotion benchmarking experiments use the standard IsaacLab velocity-conditioned locomotion environments with default reward weights.
The following reward terms are shared across all robot configurations:

\begin{itemize}
    \item \textbf{Linear velocity tracking}: Exponential reward for matching the commanded linear velocity in the horizontal plane.
    \item \textbf{Angular velocity tracking}: Exponential reward for matching the commanded yaw rate.
    \item \textbf{Feet air time}: Reward for keeping feet in the air for a sufficient duration during each step.
    \item \textbf{Angular velocity (xy)}: Penalty for roll and pitch angular velocities.
    \item \textbf{Joint accelerations}: Penalty for large joint accelerations.
    \item \textbf{Action rate}: Penalty for rapid changes in actions between consecutive timesteps.
\end{itemize}

The default quadruped environments (Go2, \revision{Spot}) use less reward shaping, relying on velocity tracking and a weak air time reward (weight: 0.125) to allow gaits to emerge naturally. These environments also include penalties for vertical base velocity, joint torques, and undesired contacts on non-foot body parts (e.g., thighs).

The Spot quadruped environment introduces several additional terms: a {gait synchronization} reward that encourages diagonal foot pair coordination for trotting, a {foot clearance} reward for achieving target swing height, an {air time variance} penalty for consistent swing timing across legs, and explicit {foot slip} and {body orientation} penalties.

The humanoid environments (H1, G1) include biped-specific adaptations: a modified air time reward for two-legged locomotion, {joint deviation} penalties that regularize hip, arm, and torso positions toward default poses, {feet slide} penalties, and a large {termination penalty} to discourage falling. The G1 configuration uses higher angular velocity tracking weight and enables joint torque penalties, while H1 disables both torque penalties and vertical velocity penalties. Neither humanoid environment includes gait synchronization rewards.

For more details, we refer to the open-source IsaacLab~\citep{mittal2023orbit} repository.

\subsection{Motion Tracking}

Our motion tracking experiments train a policy for the Unitree G1 robot, which has 29 degrees of freedom (DoF). The control frequency is set to 50 Hz (with a simulation timestep $\Delta t = 0.005$ s and decimation of 4). The reward design, observation space, and termination conditions follow BeyondMimic~\citep{liao2026beyondmimic}. The following sections detail the specific hyperparameters used.

\begin{table*}[t!]
    \centering
    \begin{tabular}{@{}lcl@{}}
        \toprule
        Hyperparameter category & Used value & Sweep range and notes \\
        \midrule
        \multicolumn{3}{l}{\textit{FPO++}} \\
        \midrule
        Flow integration steps & $64$ & $\{8, 16, 32, 64\}$ \\
        Network output & $u$ & $\{u,x_0\}$, $x_0$ denotes data \\
        Samples per action & $16$ & $\{8, 16, 32\}$ \\
        \midrule
        \multicolumn{3}{l}{\textit{Training}} \\
        \midrule
        Learning rate & $1\times10^{-4}$ & $1\times10^{-5}$, $1\times10^{-4}$, $3\times10^{-4}$ \\
        Weight decay / Adam betas & $1\times10^{-4}$ / $(0.9, 0.95)$ & AdamW parameters \\
        Clip parameter & $0.05$ & $\{0.03, 0.04, 0.05, 0.06\}$ \\
        Discount factor ($\gamma$) & $0.99$ & \\
        GAE lambda ($\lambda$) & $0.95$ & \\
        Learning epochs & $16$ for Go2, others $32$ & \\
        Minibatches per update & $4$ \\
        Running observation normalization & Yes \\
        \bottomrule
    \end{tabular}
    \caption{\textbf{FPO++ and training hyperparameters used for locomotion.}
    We report both the final values we use for experiments and the values we performed sweeps over.
    }
    \label{tab:flow_rsl_rl_hyperparameters_locomotion}
\end{table*}

\begin{table*}[h!]
    \centering
    \begin{tabular}{@{}lcl@{}}
        \toprule
        Hyperparameter category & Used value & Sweep range and notes \\
        \midrule
        \multicolumn{3}{l}{\textit{FPO++}} \\
        \midrule
        Flow integration steps & $50$ & $\{10, 50\}$ \\
        Network output & $u$ & $\{u,x_0\}$, $x_0$ denotes data \\
        Samples per action & $16$ & $\{8, 16, 32\}$ \\
        \midrule
        \multicolumn{3}{l}{\textit{Training}} \\
        \midrule
        Learning rate & $3\times10^{-4}$ & \\
        Weight decay / Adam betas & $1\times10^{-4}$ / $(0.9, 0.95)$ & AdamW parameters \\
        Clip parameter & $0.01$ & $\{0.01, 0.05, 0.1\}$ \\
        Discount factor ($\gamma$) & $0.99$ & \\
        GAE lambda ($\lambda$) & $0.95$ & \\
        Learning epochs & $5$ & \\
        Minibatches per update & $4$ \\
        Running observation normalization & Yes \\
        \bottomrule
    \end{tabular}
    \caption{
        \textbf{FPO++ and training hyperparameters used for motion tracking.}
        We report both the final values we use for experiments and the values we performed sweeps over.
    }
    \label{tab:flow_rsl_rl_hyperparameters_motion}
\end{table*}

\subsubsection{Domain Randomization}
Domain randomization is applied to improve the policy's sim-to-real transfer ability:
\begin{itemize}
    \item \textbf{Physics Material}: Static friction (Uniform $[0.3, 1.6]$), Dynamic friction (Uniform $[0.3, 1.2]$), and Restitution (Uniform $[0.0, 0.5]$) are sampled at startup.
    \item \textbf{Joint Defaults}: Default joint angles are uniformly offset by $\mathcal{U}(-0.01, 0.01)$ rad at startup.
    \item \textbf{Center of Mass (COM)}: The torso COM is offset uniformly in ($x, y, z$) at startup.
    \item \textbf{External Forces}: Pushes are applied at regular intervals of 2.0 to 3.0 seconds. Each push applies a random linear velocity between -0.5 and 0.5 meters per second along the forward and sideways directions, and between -0.2 and 0.2 meters per second in the vertical direction. Angular velocity is randomized between -0.52 and 0.52 radians per second for pitch and roll, and between -0.78 and 0.78 radians per second for yaw.
    \item \textbf{Actuator Command Delay}: Actuator latency is simulated by applying a random delay of $0$ to $2$ simulation steps (corresponding to $0$ ms to $10$ ms at $\Delta t = 0.005$ s) to the control commands (joint positions, velocities, efforts) at each environment reset to improve robustness and sim-to-real transfer.
    \item \textbf{Motion Initialization}: Root position, orientation, and joint angle offsets are applied uniformly at episode reset.
\end{itemize}

\subsubsection{Training Hyperparameters}
Training is conducted across 4096 parallel environments, with a rollout length of 96 steps per environment. Both the actor and critic networks use 3-layer MLPs with hidden \revision{unit} sizes (1024, 512, 256). We will release code for further details.
The Booster T1 locomotion policy uses the same hyperparameters as the IsaacLab locomotion experiments.
We train motion tracking policies with 50 flow integration steps; at deployment, we use zero-sampling with 5 flow integration steps to reduce latency.
Additional hyperparameters can be found in Table~\ref{tab:flow_rsl_rl_hyperparameters_motion}.

\subsubsection{CFM loss implementation details}
\label{app:cfm_loss_alternatives}

\begin{figure*}[t!]
    \centering
    \subfloat[Without CFM loss clamp.\label{fig:huber_no_inner}]{%
        \includegraphics[width=0.8\textwidth]{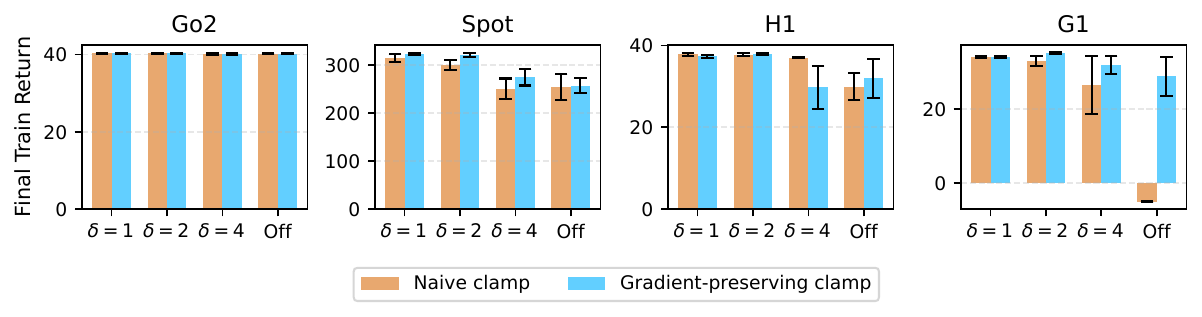}%
    }
    
    \vspace{1em}
    
    \subfloat[With CFM loss clamp.\label{fig:huber_with_inner}]{%
        \includegraphics[width=0.8\textwidth]{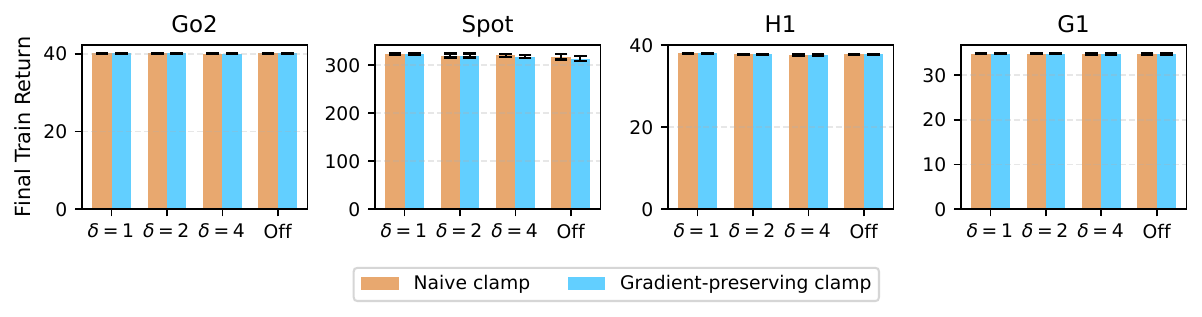}%
    }

    \caption{
        \textbf{Clamping CFM losses improves stability.}
        The x-axis varies the Huber threshold $\delta$; ``Off'' uses squared CFM loss without Huber.
        Loss-difference clamping is enabled throughout: orange uses naive clamping, and blue preserves gradients through the clamp.
        Panels (a) and (b) respectively disable and enable clamping of the old- and new-policy CFM losses before subtraction.
        Bars show mean final training return with standard error over five seeds.
    }
    \label{fig:appendix_huber_ablation}
\end{figure*}

The FPO and FPO++ ratios (Equation~\ref{eq:fpo-ratio}) both exponentiate a difference between squared CFM losses.
An early concern we had was that exponentiating a value computed from squares could cause numerical instabilities for outlier epsilon values.
We ran experiments with details like Huber instead of squared CFM losses, as well as by applying both naive and gradient-preserving clamp operations to CFM loss differences (Figure~\ref{fig:appendix_huber_ablation}).
We found that a simple approach of (i) clamping CFM losses before taking differences and (ii) then clamping the difference before exponentiation was sufficient for stabilizing training.

\subsection{Manipulation finetuning}

We use actor learning rates of $1\times10^{-5}$ and critic learning rates of $1\times10^{-4}$ in the final manipulation results. The generalized advantage estimation (GAE) parameter is fixed to $\lambda = 0.99$ across all five manipulation tasks. The discount factor $\gamma$ varies by task: $\gamma=0.99$ for \emph{RoboMimic Can}, $\gamma=0.995$ for \emph{RoboMimic Square} and \emph{DexMimicGen Box Cleanup}, and $\gamma=0.999$ for \emph{DexMimicGen Tray Lift} and \emph{Threading}. These hyperparameters are shared across vanilla FPO and DPPO variants in Figure~\ref{fig:manipulation_comparison}. For both FPO++ and vanilla FPO, we use 10 flow sampling steps and fix the number of Monte Carlo samples per action chunk to 8 across all manipulation tasks. Policies sample action chunks with horizon length 16, and during \revision{finetuning} we compute chunk-level ratios by summing CFM losses across all chunk timesteps.

For our comparisons between FPO++, vanilla FPO, and two DPPO~\citep{ren2024diffusion} variants in Figure~\ref{fig:manipulation_comparison}, we adapt DPPO implementations from~\citep{mcallister2025flow}.
Some implementation details differ from the original DPPO implementation, for example, adopting velocity prediction instead of epsilon prediction~\citep{ho2020denoising}.
This allowed comparisons between methods to be initialized from the same base policy and evaluated under more similar conditions.
We extensively tuned baselines for fairness, sweeping over hyperparameters including PPO clipping thresholds, gradient norm clipping, and exploration noise scales.
We present additional baselines in Appendix~\ref{app:manipulation_more_baselines}.

\begin{figure*}[h!]
    \centering
    \includegraphics[width=\textwidth]{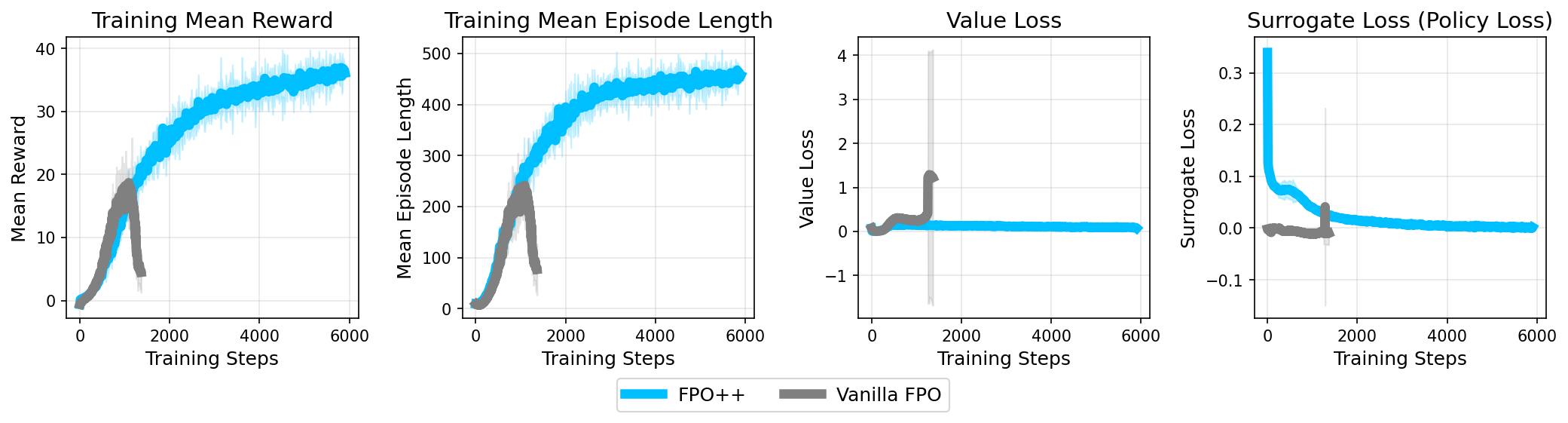}
    \vspace{-1.5em}
    \caption{
        \textbf{Motion tracking training curves.} Vanilla FPO fails in learning to track the motion, while FPO++ trains stably.
        Hyperparameters are shared between both runs.
    }
    \label{fig:motion_tracking_fpo_improved}
\end{figure*}

\section{More experiments}

\subsection{Sample efficiency and emergent gaits versus Gaussian PPO}

\begin{figure}[t!]
    \centering
    \includegraphics[width=0.8\linewidth]{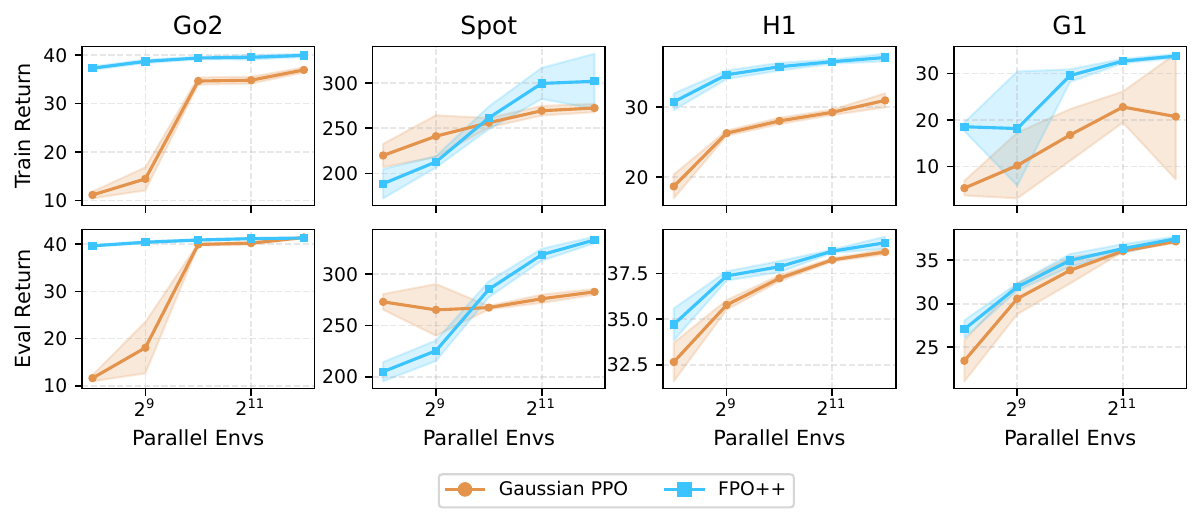}
    \caption{
        \textbf{Sample efficiency for FPO++ and our Gaussian PPO baseline.}
        We compare policy returns between Gaussian PPO and FPO++ with environment counts $\in \{2^8,2^9,2^{10},2^{11},2^{12}\}$ after 1500 updates for quadrupeds and 2000 updates for humanoids.
        Minibatch size per policy update increases linearly with the number of parallel environments; other hyperparameters are fixed between runs.
        Filled regions show standard deviation computed over 5 seeds.
    }
    \label{fig:ppo_compare_numenvs}
\end{figure}

\begin{figure*}[t!]
    \centering
    \includegraphics[width=0.8\textwidth]{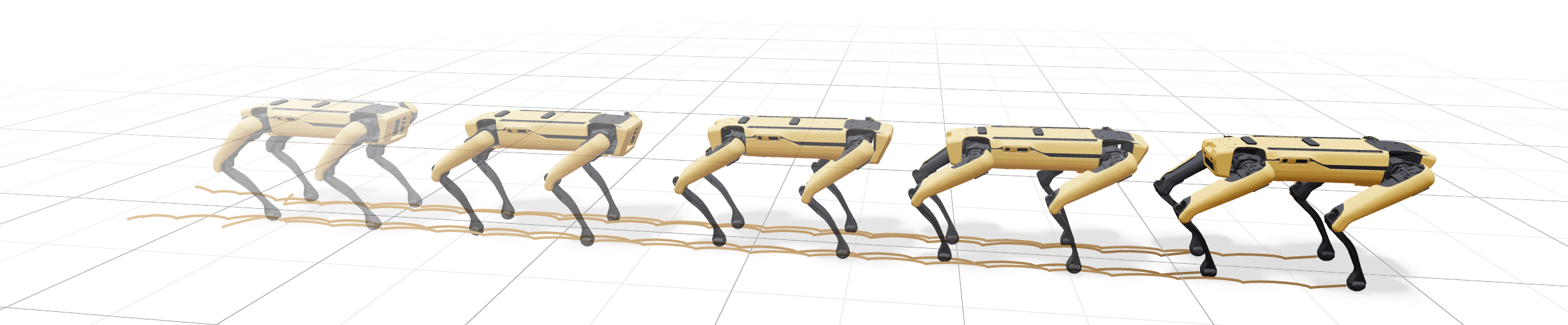}
    \includegraphics[width=0.8\textwidth]{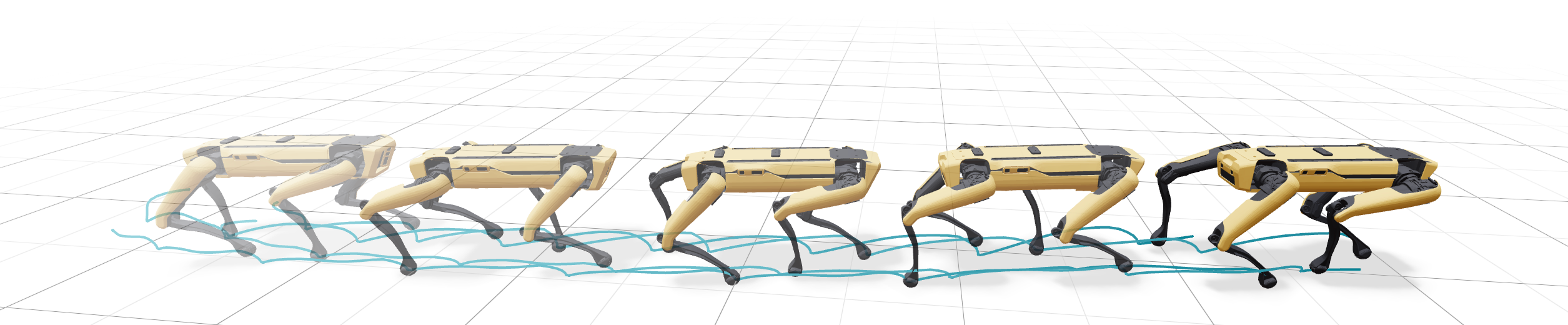}
    \vspace{1em}
    \caption{
        \textbf{Different gaits from different algorithms.}
        Rollouts using Gaussian PPO (top) and FPO++ (bottom) for 1500 policy updates.
        Policies are trained with the same rewards, 4096 parallel environments, and given the same forward velocity command (\revision{1~m/s}).
        We found differences reproducible across seeds, clip parameters, and learning rates.
        We use Viser~\citep{yi2025viser} for visualization.
    }
    \label{fig:ppo_fpo_spot_gait}
\end{figure*}

Figure~\ref{fig:ppo_compare_numenvs} compares policy returns between Gaussian PPO and FPO++ across locomotion tasks as we vary the number of parallel environments.
Figure~\ref{fig:ppo_fpo_spot_gait} shows the qualitatively different gaits that the two algorithms converge to under identical rewards.

\subsection{Comparison with vanilla FPO for motion tracking}

Figure \ref{fig:motion_tracking_fpo_improved} shows the training curves for the motion-tracking policy that we deploy on the real G1 robot. As the plots illustrate, vanilla FPO initially learns but quickly collapses: the mean reward and episode length peak early and then deteriorate, while both the value loss and surrogate loss exhibit large spikes, indicating numerical instability. In contrast, FPO++ maintains stable value and policy losses throughout, continues improving monotonically, and successfully converges to the high-return policy used for real-world deployment. These curves provide evidence for a core motivation of our work: vanilla FPO is unstable on realistic, high-DoF robot control tasks, whereas the algorithmic components of FPO++ prevent collapse and enable successful sim-to-real transfer. We will release training code for reproducibility.

\subsection{\revision{Comparison with Gaussian PPO for motion tracking}}
\label{app:gaussian_motion_tracking}

\begin{figure}[!htbp]
    \centering
    
    \subfloat[\revision{Without entropy regularization or adaptive learning rate.}\label{fig:motion_tracking_comp_no_ent}]{%
        \includegraphics[width=0.49\linewidth]{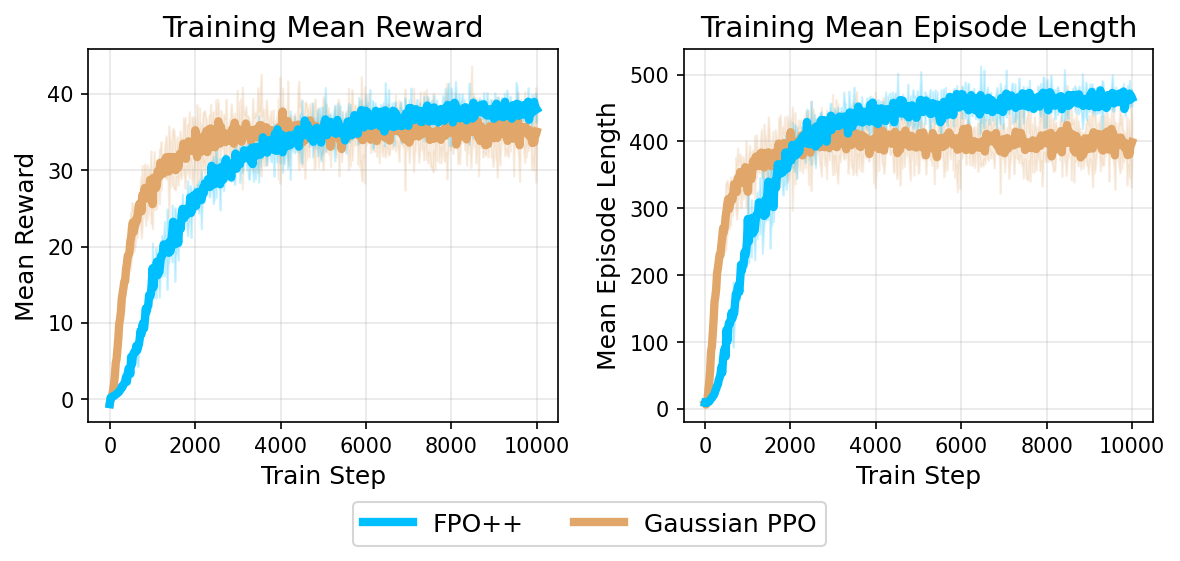}%
    }\hfill
    \subfloat[\revision{With entropy regularization and adaptive learning rate.}\label{fig:motion_tracking_comp_yes_ent}]{%
        \includegraphics[width=0.49\linewidth]{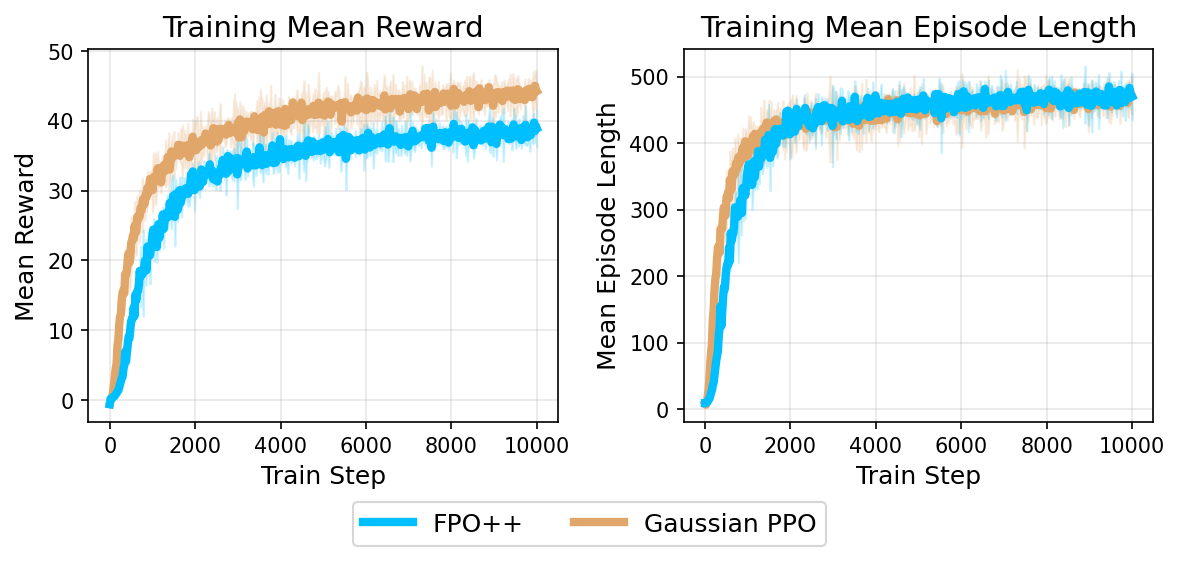}%
    }
    \caption{\textbf{Training curves compared to Gaussian PPO for motion tracking.} Blue denotes FPO++ and orange Gaussian PPO. Without entropy regularization or adaptive learning rates, FPO++ achieves higher final returns and longer episodes. With both components enabled, Gaussian PPO attains higher returns, while episode lengths are similar.}
    \label{fig:motion_tracking_comparison}
\end{figure}

{
FPO++ enables successful sim-to-real transfer for high-DoF motion tracking, but this does not imply higher simulation returns than tuned Gaussian PPO. Figure~\ref{fig:motion_tracking_comparison} compares the two algorithms with and without entropy regularization and adaptive learning rates. Without these components, FPO++ achieves higher final returns and longer episodes than the simplified Gaussian PPO baseline. With these components enabled for both algorithms, Gaussian PPO achieves higher returns, while FPO++ attains similar or slightly longer episodes.

For this comparison, we augment FPO++ with simple entropy regularization and adaptive learning-rate mechanisms. These are experimental variants rather than changes to the core FPO++ objective.

\textbf{Entropy regularization.}
We use a non-parametric Kozachenko--Leonenko estimator based on nearest-neighbor distances between action predictions. This estimates local density using pairwise $L_2$ distances and supplies an entropy bonus to encourage exploration.

\textbf{Adaptive learning rate.}
The FPO++ variant uses an $L_2$-based noise-prediction-change proxy to adapt the learning rate. Noise predictions are recovered algebraically from velocity predictions and sampled actions; the resulting statistic measures policy change without explicitly computing action likelihoods. It is a heuristic proxy, not an exact policy KL divergence.

These components improve FPO++ training in this comparison but do not close the return gap to tuned Gaussian PPO. More effective regularization and learning-rate adaptation for flow policies remain directions for future work.
}

\subsection{Further comparisons with baselines in manipulation finetuning}
\label{app:manipulation_more_baselines}

\begin{figure*}[t!]
    \centering
    \includegraphics[width=0.7\textwidth]{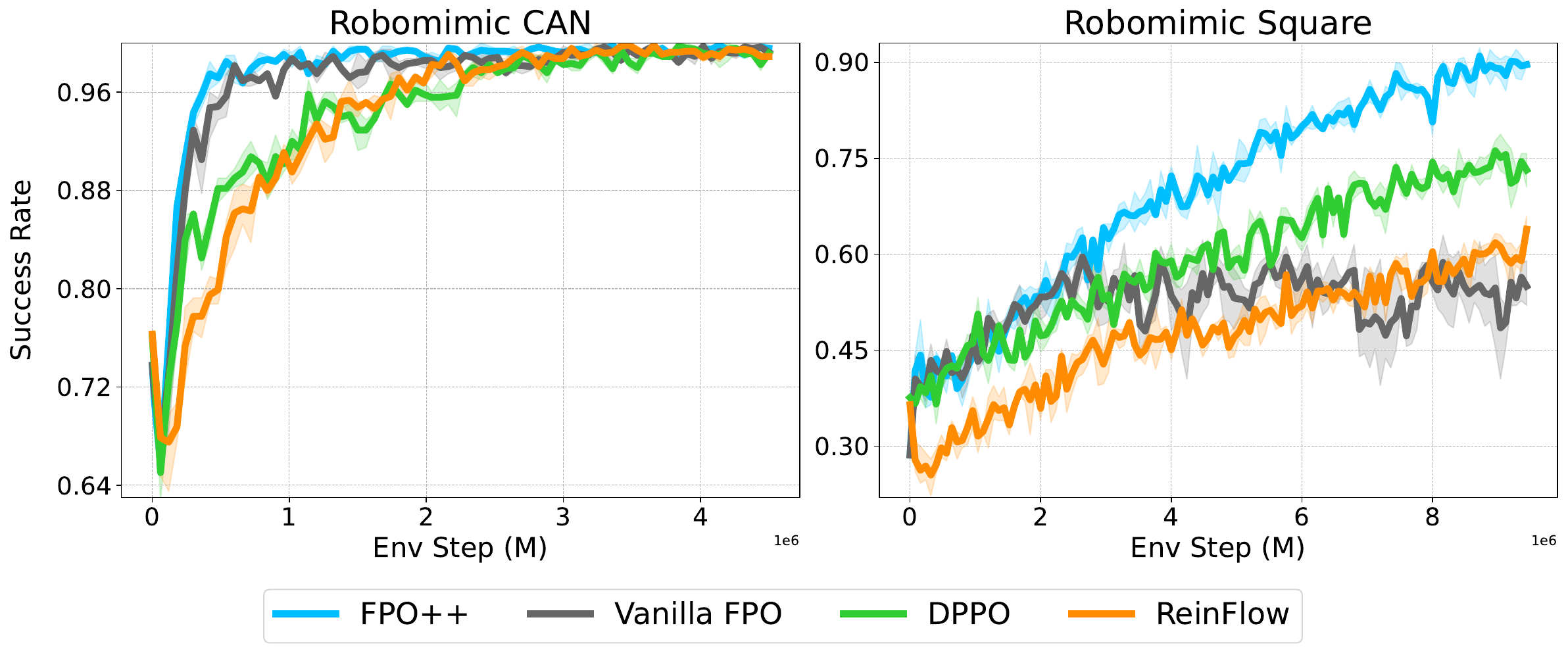}
    \caption{
    \textbf{Additional Manipulation Comparison.}
    \revision{Finetuning} success rates on the RoboMimic \emph{Can} and \emph{Square} tasks for flow-based RL methods.
    All methods start from \revision{pretrained} behavior-cloning base policies; FPO++ and \revision{vanilla} FPO share the same base policy, while DPPO and ReinFlow use architectures and implementation from their original papers. Please refer to the base policy setting written in the text for further details.
    FPO++ learns fastest and attains the highest final success on both tasks, while \revision{vanilla} FPO, DPPO, and ReinFlow remain stable but underperform relative to FPO++.
}
\label{fig:manipulation_comparison_v2}
\end{figure*}

We compare FPO++ with DPPO and ReinFlow~\citep{zhang2025reinflow} using the original implementations of these methods in Figure~\ref{fig:manipulation_comparison_v2}.
In this setting, we adapt our training pipeline to match their implementation details, including adopting an action chunk size of four, rather than sixteen, to align with the DPPO/ReinFlow formulation. We train FPO++ and vanilla FPO within this adapted codebase, and retrain DPPO and ReinFlow using the authors' released code, using comparable base policies and evaluating all methods under the same simulation environments. This additional experiment allows for a further algorithmic comparison while controlling for implementation-specific factors.

\textbf{Base policy setting.}
The base policies' performance, reflected in the success rates at $0$ total environment steps in Figure \ref{fig:manipulation_comparison}, provides the anchor for \revision{finetuning}. FPO++ and \revision{vanilla} FPO share the same base policy, while DPPO and ReinFlow utilize different \revision{pretrained} policies structured according to their original papers. The policies for DPPO and ReinFlow were trained and evaluated using $100$ denoising/flow steps, whereas the base policy for FPO++ and \revision{vanilla} FPO was trained and evaluated with $10$ flow steps. The base policy success rates (evaluated on 1,000 episodes) are
\begin{itemize}
    \item \textbf{Can task}: \revision{FPO++/vanilla FPO} (73.76\%), ReinFlow (\revision{76.30\%}), and DPPO (\revision{76.10\%}).
    \item \textbf{Square task}: DPPO (\revision{37.60\%}), ReinFlow (\revision{36.50\%}), and \revision{FPO++/vanilla FPO} (28.62\%).
\end{itemize}

\textbf{Analysis of \revision{finetuning} performance.}
Figure~\ref{fig:manipulation_comparison_v2} shows \revision{finetuning} success rates, evaluated by collecting 200 episodes per checkpoint using 50 parallel simulation environments. FPO++ achieves the highest final success rates on both the Can and Square tasks. On Can, FPO++ demonstrates rapid early learning, reaching high success significantly faster than all baselines. Notably, vanilla FPO also performs robustly in this \revision{finetuning} setting, achieving competitive performance without the policy collapse or gradual degradation as observed in Figure~\ref{fig:manipulation_comparison}. Nevertheless, FPO++ yields substantial gains over vanilla FPO, both in terms of convergence speed and final performance on the more challenging Square task, which requires higher-precision control. 

\begin{figure*}[t!]
    \centering
    \includegraphics[width=0.96\textwidth]{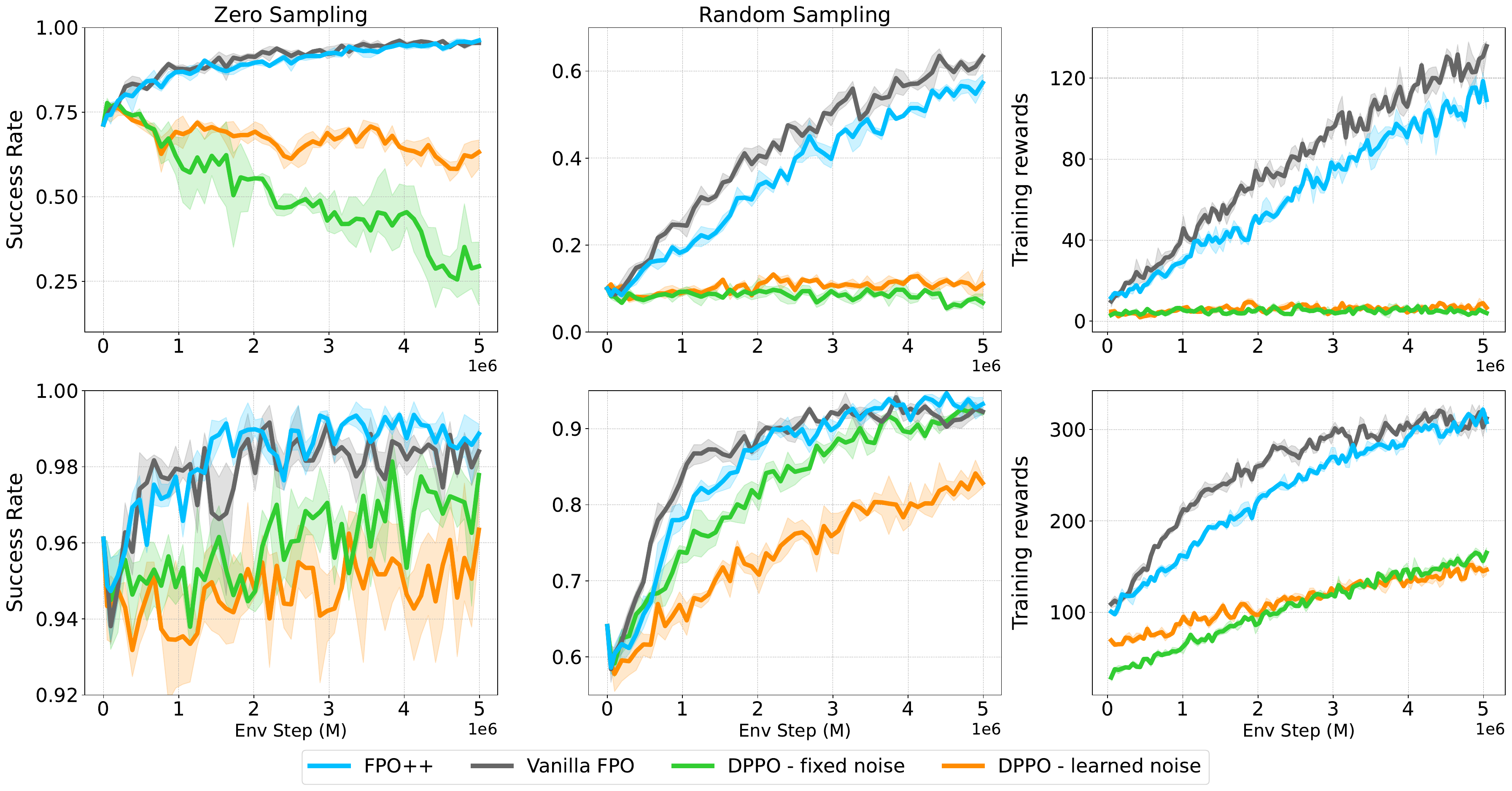}
    \caption{\textbf{Analysis of base policy quality in RoboMimic Can.} This figure illustrates how the initial success rate of the base policy impacts different \revision{finetuning} methods. The top row shows results for a base policy with 71.35\% success under zero-sampling versus 10.00\% under random sampling, while the bottom row depicts a higher-quality base policy (96.11\% zero-sampling / 64.36\% random sampling), demonstrating that DPPO variants are significantly more sensitive to the base policy quality.
}
\label{fig:manipulation_can_details}
\end{figure*}

\begin{figure}[t!]
    \centering
    \includegraphics[width=0.8\linewidth]{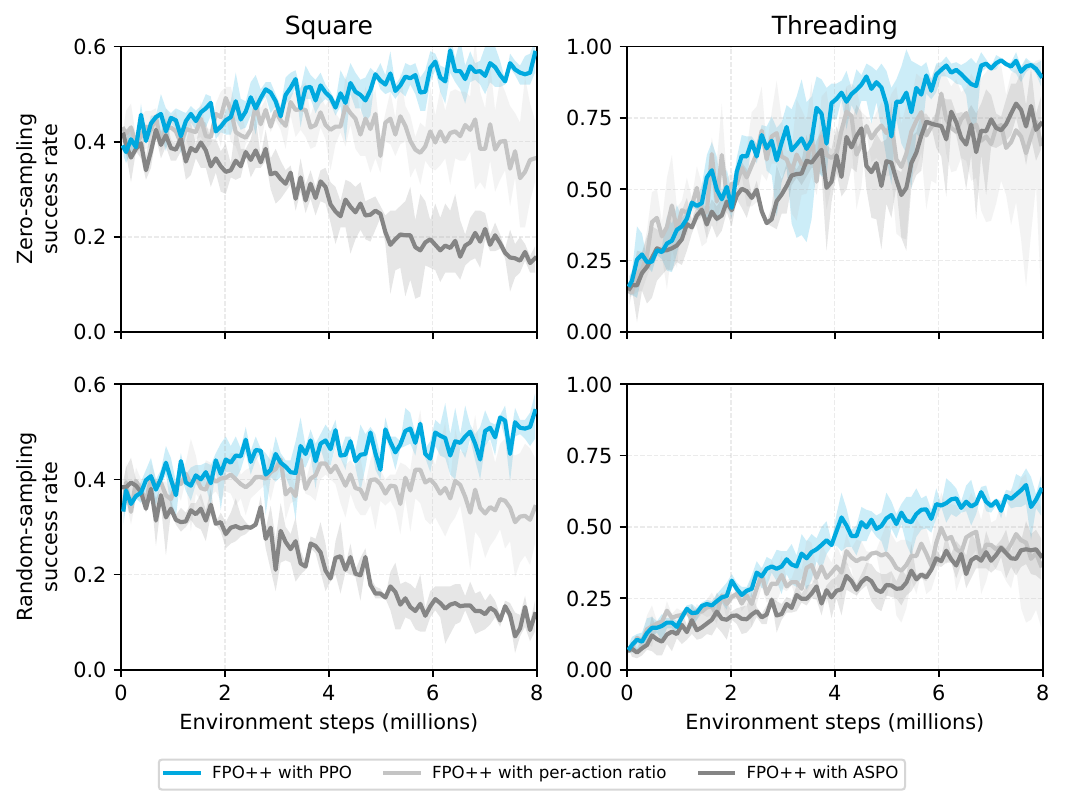}
    
    \caption{\textbf{Ablation study in manipulation tasks.} Blue denotes per-sample ratios with PPO, light gray per-action ratios with PPO, and dark gray per-sample ratios with ASPO. Results are averaged over three training seeds per configuration, with 200 evaluation episodes per checkpoint per seed. Per-sample PPO achieves the highest final mean success on both Square and Threading. ASPO underperforms per-sample PPO in both tasks, despite its benefits for locomotion.
}
\label{fig:manipulation_ablation}
\end{figure}

\subsection{Detailed analysis on the RoboMimic Can experiment}
\label{app:can_success_rate_analysis}

Although the DPPO paper reports high success rates on the Can task, our experiments in the main text show DPPO variants struggling. We attribute this discrepancy to differences between base policies.

As illustrated in Figure~\ref{fig:manipulation_can_details}, our flow-based base policy exhibits a high success rate when evaluated with zero-sampling. However, the success rate under the random sampling used for exploration during training is significantly lower. As a result, very few trajectories successfully reach the goal in each rollout phase. Combined with rewards only provided at the end of rollouts, this results in high gradient variance.

We find that FPO++ and vanilla FPO are relatively robust to this high-variance learning regime, while DPPO variants degrade substantially. In particular, although all methods begin from the same base policy and therefore share the same initial success rate under random sampling, FPO-based methods achieve noticeably higher training returns early in \revision{finetuning}, providing sufficient reward signal for stable policy-gradient updates. In contrast, DPPO introduces additional stochasticity by injecting noise at each diffusion timestep during training, which further reduces the probability that rollouts reach the task completion state and leads to near-zero returns for a large fraction of trajectories. This exacerbates gradient noise and prevents effective learning in the early stages of training.
As shown in the bottom row of Figure~\ref{fig:manipulation_can_details}, when the base policy's random sampling success rate is high enough (\revision{64.36\%}) to provide consistent initial rewards, DPPO variants succeed in \revision{finetuning} the policy and achieving competitive success rates.

\subsection{Ablation study for manipulation finetuning}
\label{app:manipulation_ablation}
In our main manipulation \revision{finetuning} experiments, we observed that while FPO++ consistently outperforms baselines, specific algorithmic components contribute differently to success than they do in the from-scratch locomotion setting. 

We conducted an ablation study across two manipulation tasks: the RoboMimic \textbf{Square task} and the DexMimicGen \textbf{Threading task}. These tasks both involve insertion, requiring high-precision control, and demonstrated a meaningful performance margin between FPO++ and vanilla FPO. As illustrated in Figure~\ref{fig:manipulation_ablation}, while the per-sample ratio consistently benefits FPO++ as it does in locomotion, the ASPO trust region is detrimental for manipulation \revision{finetuning}.

\revision{Appendix~\ref{app:lift_reward_regime} tests whether this stems from pretraining by repeating the PPO and ASPO comparison on Lift with randomly initialized action networks.}

\subsection{\revision{ASPO under sparse and dense rewards}}
\label{app:lift_reward_regime}
\begin{figure}[!htbp]
    \centering
    
    \includegraphics[width=\textwidth]{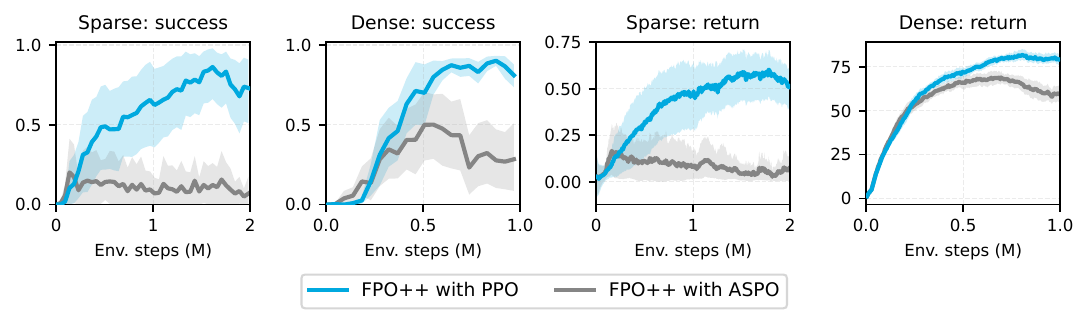}
    \caption{\textbf{ASPO sensitivity to the reward regime in Lift.} PPO and ASPO each use their best-tuned configuration for each reward setting, evaluated over ten training seeds. Success uses zero-sampling and the task-completion criterion.}
    \label{fig:lift_reward_regime}
\end{figure}

{
\textbf{Experiment setup.}
To test whether ASPO's limitation is specific to \revision{finetuning}, we compare PPO and ASPO trust regions in FPO++ on RoboMimic Lift, using the best-performing hyperparameter configuration found for each trust region under each reward setting. Action networks are randomly initialized, while the shared pretrained vision encoder remains frozen. We use ten training seeds per configuration, with budgets of 2M steps for sparse binary rewards and 1M for dense shaped rewards. Each checkpoint is evaluated over 100 episodes per seed using the same task-completion criterion in both settings.

\textbf{Results.}
PPO achieves higher final zero-sampling success under both reward settings (Figure~\ref{fig:lift_reward_regime}): 0.73 versus 0.07 for ASPO with sparse rewards, and 0.81 versus 0.28 with dense rewards. Sparse ASPO barely learns, whereas dense ASPO initially improves before success declines despite relatively high shaped return. Rollouts reveal grasp-and-hover behavior that earns shaping reward without completing the lift.

\textbf{Analysis.}
These results suggest that initialization alone does not explain ASPO's limitations, even when each trust region uses its best-tuned configuration. With sparse rewards, limiting likelihood decreases for negative-advantage actions may keep the policy near its random initialization, hindering bootstrapping from rare successes. With dense rewards, the divergence between success and return highlights the importance of alignment with task completion. This supports the reward-dependent guidance in Section~\ref{sec:exp_ablations}.
}

\section{Analysis of training dynamics}

\subsection{Empirical gradient variance}
\label{app:measuring_variance}

\begin{figure*}[t!]
    \centering
    \includegraphics[width=0.7\textwidth]{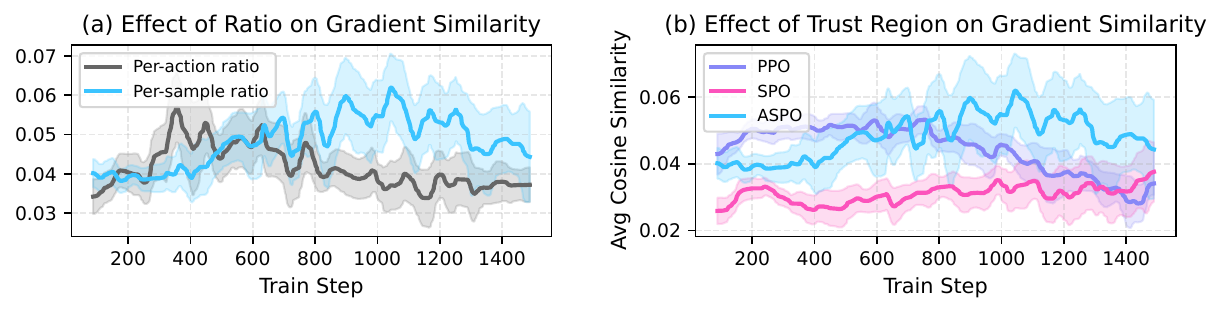}
    \caption{
        \textbf{Algorithmic updates in FPO++ reduce gradient variance.}
        We observe that the per-sample ratio and ASPO trust region result in higher cosine similarity between gradients computed within each policy update, especially during the second half of training.
        Averages and standard deviations are reported over 5 seeds.
    }
    \label{fig:gradient_variance}
\end{figure*}

One explanation for FPO++'s improved stability over FPO is reduced gradient variance: finer-grained clipping with the per-sample ratio creates a larger effective batch size, while for negative advantages, ASPO provides gradients even for ratios beyond the trust region.
We plot a cosine similarity metric inspired by~\citep{ilyas2018closer} in Figure~\ref{fig:gradient_variance}:
for each policy update, we compute cosine similarities between individual gradients and the average gradient within the policy update.
We find higher similarity when the proposed ASPO and per-sample ratios are used during training.

\subsection{Entropy and full-action-space flow fields}
\label{app:full_flow_field}

\begin{figure}[!htbp]
    \centering
    \includegraphics[width=0.5\linewidth]{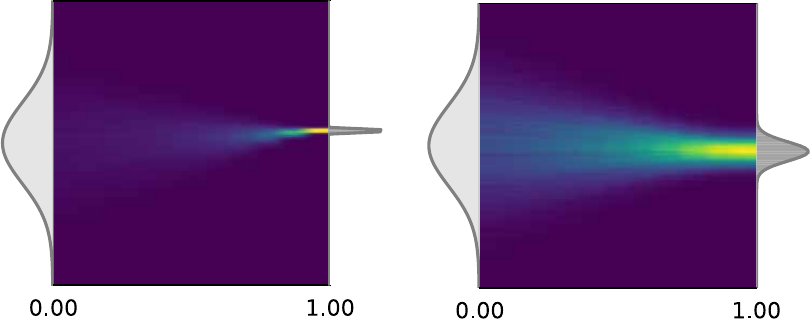}
    \caption{
        \textbf{Effect of ASPO on entropy.}
        We show a flow field during training of H1 locomotion for a single robot joint, with FPO++ using PPO clipping (left) and ASPO clipping (right).
        The x-axis is integration step $\tau$.
    }
    \label{fig:entropy_small}
\end{figure}

\begin{figure*}[!htbp]
    \centering
    \includegraphics[width=0.85\textwidth]{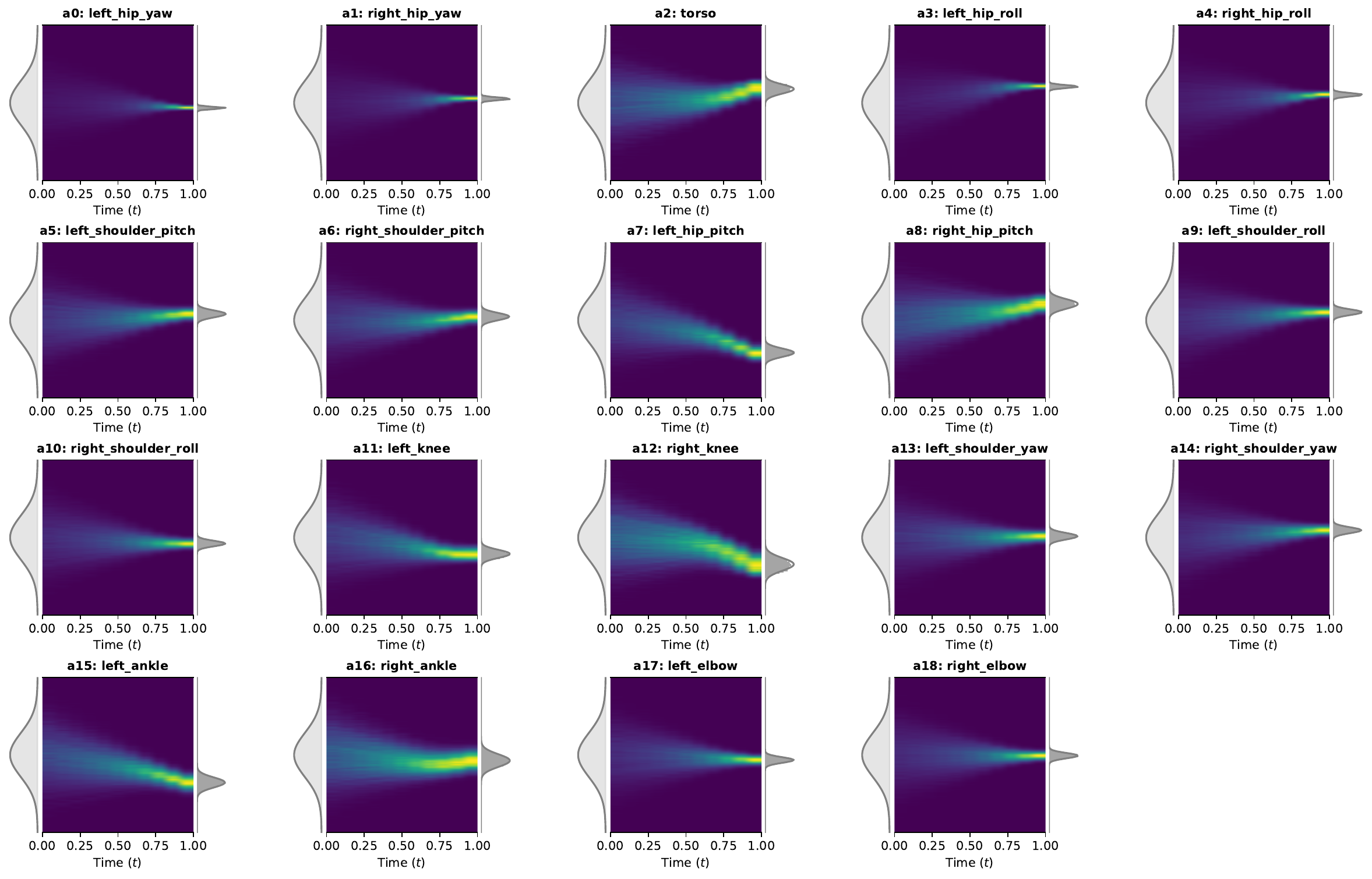}
    \caption{\textbf{FPO++ with PPO clipping at peak performance}. Policy flow field density for the Unitree H1 humanoid trained using the standard PPO trust region, captured at the point of maximum average return. The flow is still relatively broad, indicating adequate exploration.}
    \label{fig:h1_500_ppo}
\end{figure*}

\begin{figure*}[!htbp]
    \centering
    \includegraphics[width=0.85\textwidth]{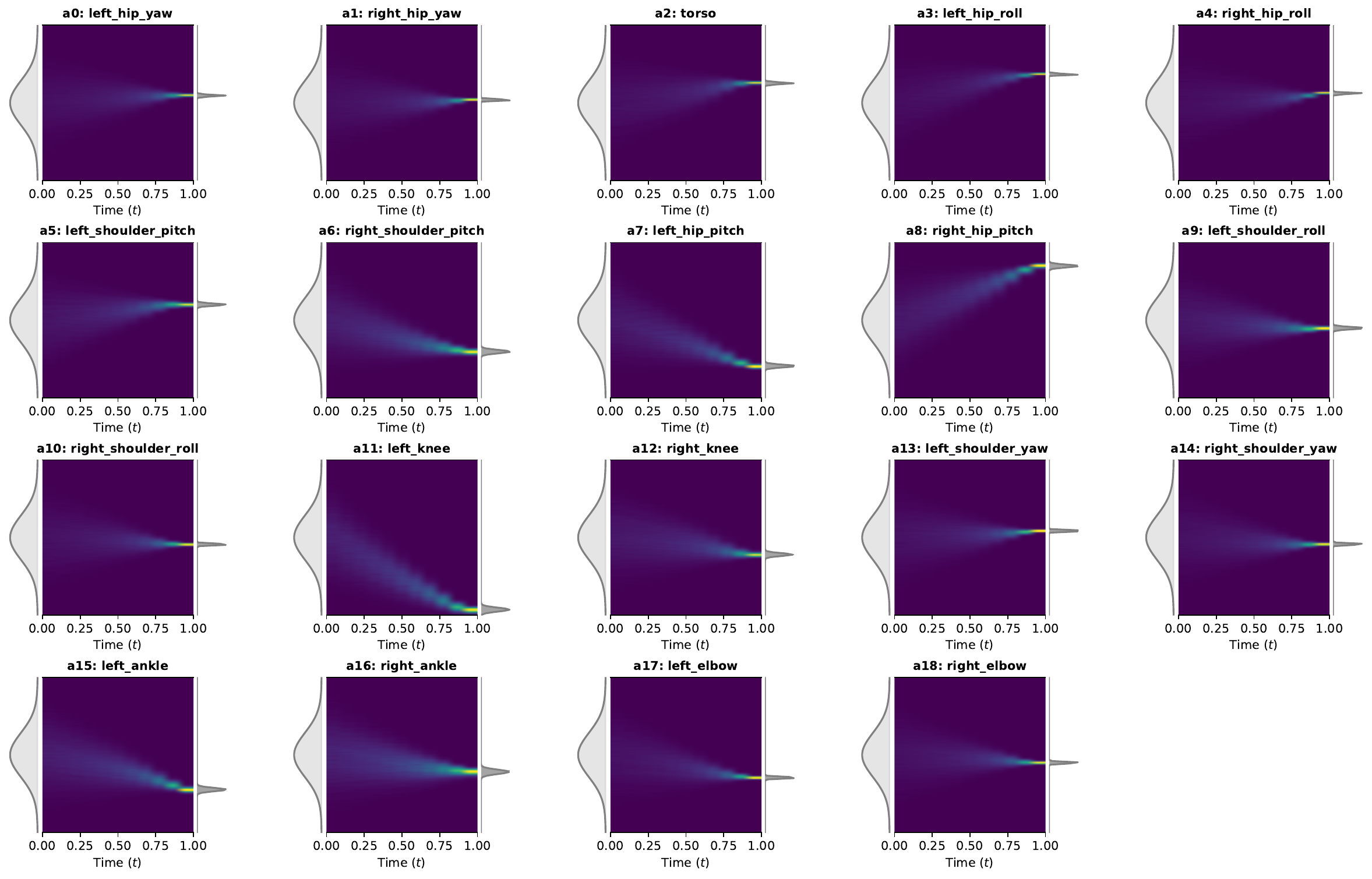}
    \caption{\textbf{FPO++ with PPO clipping during policy collapse after convergence.} Policy flow field density for the Unitree H1 humanoid trained using the standard PPO trust region, captured as the average return begins to collapse. The action distribution has narrowed significantly for many joints, demonstrating the entropy collapse that leads to instability and performance degradation.}
    \label{fig:h1_800_ppo}
\end{figure*}

\begin{figure*}[!htbp]
    \centering
    \includegraphics[width=0.85\textwidth]{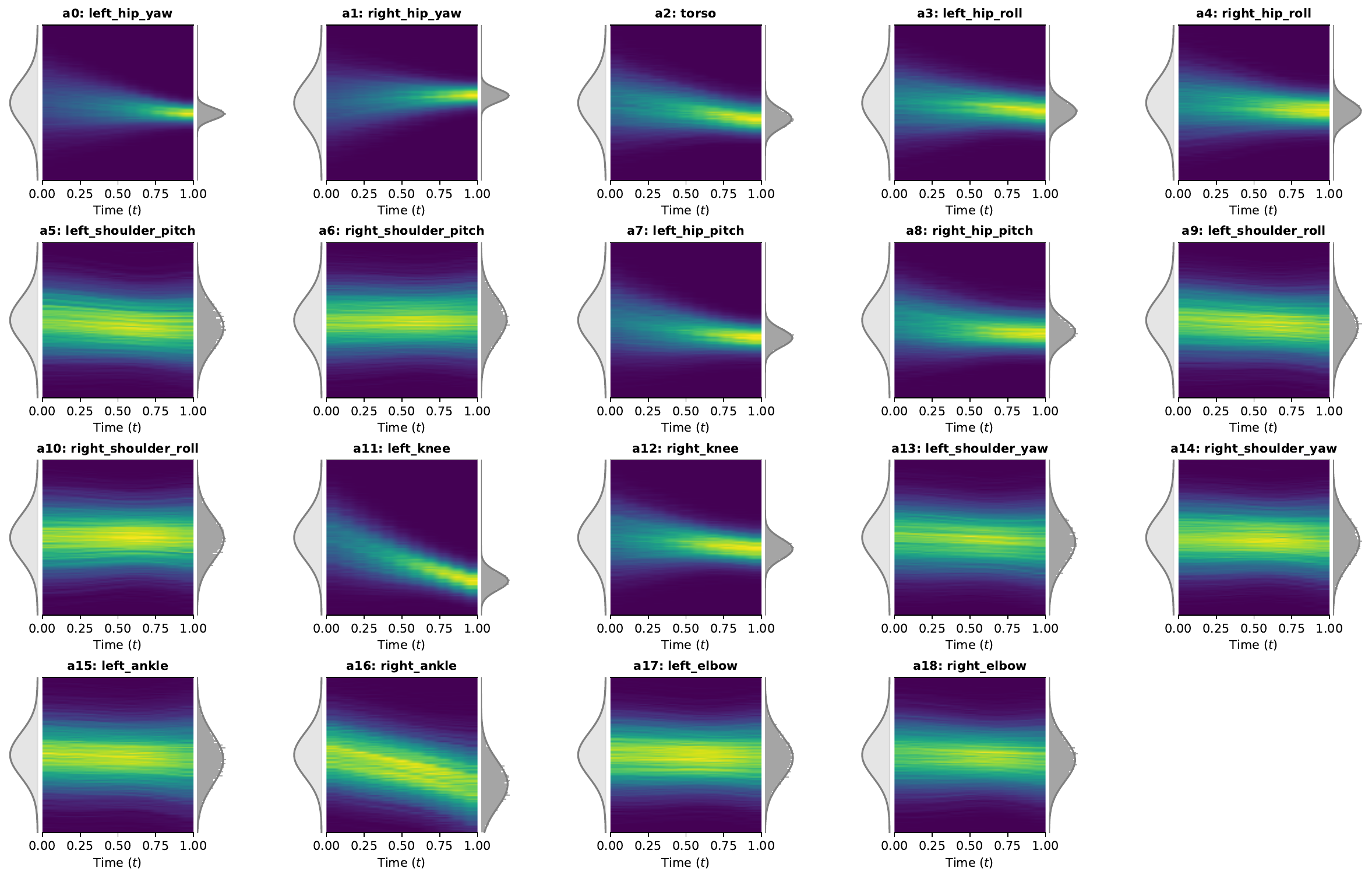}
    \caption{\textbf{FPO++ with ASPO objective at intermediate training stage.} Policy flow field density for the Unitree H1 humanoid trained using the entropy-preserving ASPO objective, captured at an intermediate stage where the average return matches the peak average return of FPO++ with the standard PPO trust region. This distribution is already wider and more exploratory than the PPO-clipped policy at a similar or later time step, contributing to stable training.}
    \label{fig:h1_600_aspo}
\end{figure*}

\begin{figure*}[!htbp]
    \centering
    \includegraphics[width=0.85\textwidth]{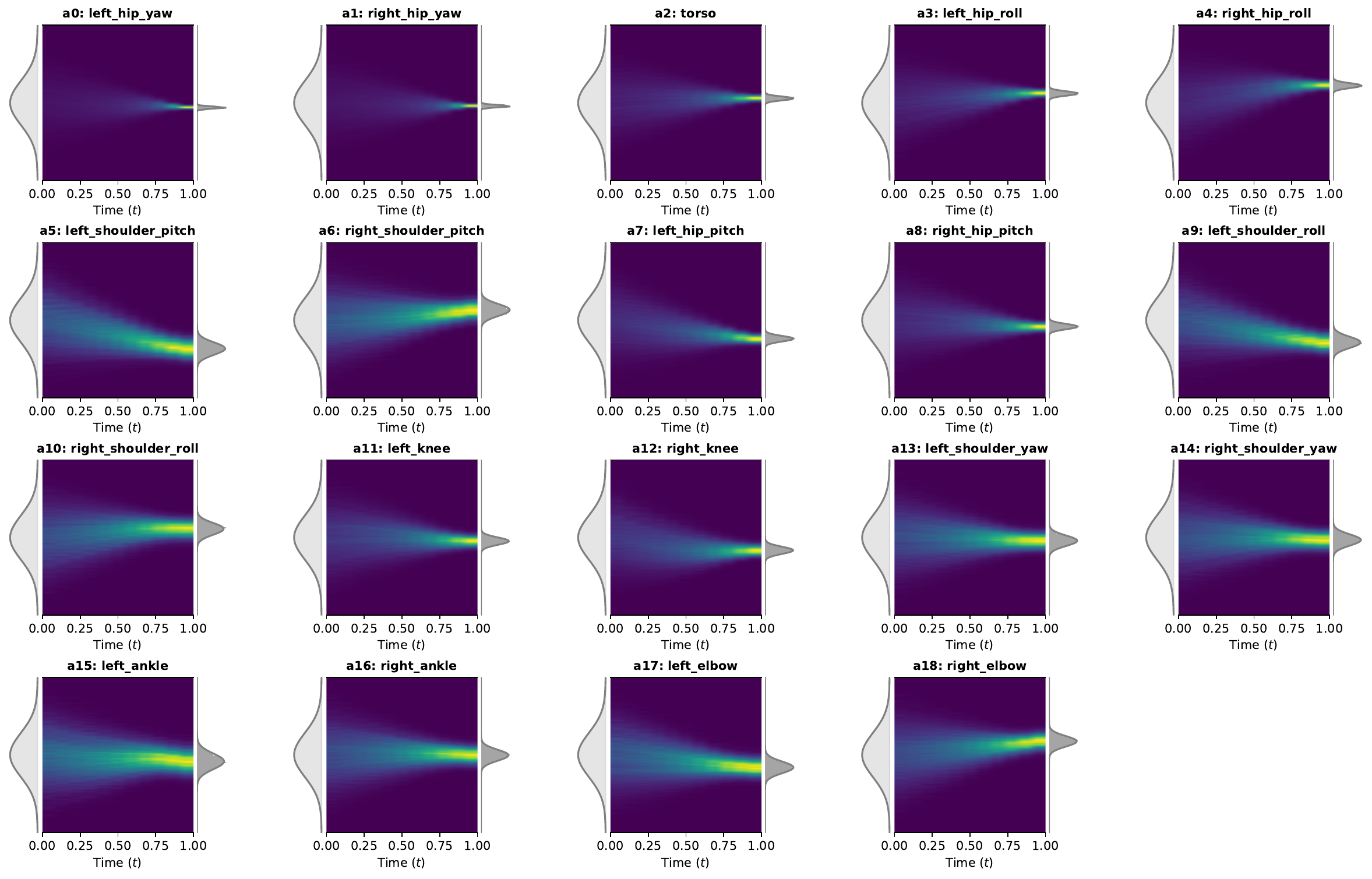}
    \caption{\textbf{FPO++ with ASPO objective at converged state.} Policy flow field density for the Unitree H1 humanoid trained using the ASPO objective, captured at its final, high-reward converged state. The distribution remains wide and exploratory, confirming that ASPO effectively preserves entropy and prevents the collapse observed with PPO clipping.}
    \label{fig:h1_1900_aspo}
\end{figure*}

In addition to the qualitative analysis presented in the main text, we provide visualizations of the flow fields across the full action space (19 DoF) for the $\text{H}1$ humanoid's velocity-conditioned locomotion task. 
These supplementary figures illustrate the impact of the trust region objective on the policy distribution at different stages of training. Specifically, the FPO++ policy trained with the standard PPO trust region exhibits a clear narrowing of the action distribution (entropy collapse) as performance degrades, evident when comparing the field at its reward peak (Figure \ref{fig:h1_500_ppo}) to the field during collapse (Figure \ref{fig:h1_800_ppo}). In contrast, the policy trained with the ASPO trust region maintains a broader, more exploratory distribution, consistently preserving entropy from an intermediate checkpoint (Figure \ref{fig:h1_600_aspo}) to its final, converged state (Figure \ref{fig:h1_1900_aspo}). This shows how the asymmetric objective prevents entropy collapse. For visualization, we plotted the transporting trajectories of the prior Gaussian noise to the action space by sampling $10{,}000$ noises per state and using $8$ discretized Euler integration steps.
Figure~\ref{fig:entropy_small} first summarizes this contrast for a single joint, comparing the flow field under PPO and ASPO clipping over the course of training; Figures~\ref{fig:h1_500_ppo}--\ref{fig:h1_1900_aspo} then show the full action space.

\FloatBarrier
\subsection{Zero-sampling}
\label{app:zero_sampling}

Policy gradient methods require stochastic policies for exploration during training.
Performance at test-time, however, can benefit from deterministically choosing actions.
During training, FPO++ policies explore by drawing initial noises from $\epsilon \sim \mathcal{N}(0, I)$ and performing Euler integration over the learned flow field; at test-time and for computing evaluation metrics, we initialize flow integration from $\epsilon = \vec{0}$.
We refer to this as \textit{zero-sampling}.
Similar to concurrent analysis on sampling in behavior cloning~\citep{pan2025much}, we find that this improves performance across tasks.
The effect can be observed in the gap between locomotion train and eval returns in Figure~\ref{fig:ppo_compare_numenvs}, in the gap between \revision{zero-sampling and random sampling} success rates for manipulation in Figure~\ref{fig:manipulation_comparison}, and in motion tracking rewards with different sampling strategies in Table~\ref{tab:zero_sampling}.

\begin{table}[!htbp]
\centering
\begin{tabular}{lcc}
\toprule
Sampling method & 5 steps & 50 steps \\
\midrule
Random sampling & $34.7 \pm 55.0$ & $38.4 \pm 26.6$ \\
Zero-sampling   & $\mathbf{45.1 \pm 27.4}$ & $\mathbf{45.5 \pm 23.2}$ \\
\bottomrule
\end{tabular}
\vspace{0.5em}
\caption{\textbf{Zero-sampling improves motion tracking.} Returns are computed over 100 rollouts of a \revision{2.5-minute} dancing sequence, with domain randomization and push perturbations. The policy is trained with 50 flow steps; we evaluate using random sampling and zero-sampling with 5 and 50 flow steps.}
\label{tab:zero_sampling}
\end{table}

\textbf{Zero-sampling is critical, especially for sim-to-real.}
The gap between rollouts with \revision{random sampling} and rollouts with zero-sampling can be drastic: G1 policies trained with $2^9$ parallel environments, for example, achieve an average train return of under 20 but eval return above 32.
Success rates in manipulation base policies, before any \revision{finetuning} is applied at all, can jump from around $10\%$ to over $70\%$.
We also observe in Table~\ref{tab:zero_sampling} that zero-sampling allows us to significantly reduce Euler integration steps with only a negligible drop in policy performance.
This enables lower-latency action sampling using the on-board computer of the robot.

\end{document}